\documentclass[lettersize,journal]{IEEEtran}
\usepackage{amsmath,amsfonts}
\usepackage{algorithm}
\usepackage{algpseudocode}
\usepackage{array}
\usepackage{subfig}
\usepackage{textcomp}
\usepackage{stfloats}
\usepackage{url}
\usepackage{verbatim}
\usepackage{graphicx}
\usepackage{cite}
\usepackage{hyperref}
\usepackage{siunitx}
\usepackage{enumitem}
\usepackage{amsthm}
\usepackage{adjustbox}
\usepackage{hhline}
\usepackage{tabularx}
\usepackage{booktabs}
\usepackage{mathtools}

\newtheorem{remark}{Remark}

\begin{document}

\title{Haptic bilateral teleoperation system for free-hand dental procedures}

\author{Lorenzo Pagliara,~\IEEEmembership{Student Member,~IEEE,}
Vincenzo Petrone,~\IEEEmembership{Member,~IEEE,}
Enrico Ferrentino,~\IEEEmembership{Member,~IEEE,}
Andrea Chiacchio, Giovanni Russo,
~\IEEEmembership{Senior Member,~IEEE}
\thanks{This research was conducted in the frame of the Department of Excellence 2023/2027 Project}
\thanks{Lorenzo Pagliara, Vincenzo Petrone, Enrico Ferrentino and Giovanni Russo are with the Department of Information Engineering, Electrical Engineering and Applied Mathematics (DIEM), University of Salerno, 84084 Fisciano, Italy,
e-mail: \{lpagliara, vipetrone, eferrentino, giovarusso\}@unisa.it}%
\thanks{Andrea Chiacchio is with the Specialization School in Oral Surgery, University of Naples Federico II, 80138  Napoli, Italy,
e-mail: andr.chiacchio@studenti.unina.it}%
\thanks{\textcopyright2026 IEEE.
Personal use of this material is permitted.
Permission from IEEE must be obtained for all other uses, in any current or future media, including reprinting/republishing this material for advertising or promotional purposes, creating new collective works, for resale or redistribution to servers or lists, or reuse of any copyrighted component of this work in other works.}
\thanks{(\textit{Corresponding author: Lorenzo Pagliara})}}

\markboth{\MakeUppercase{ACCEPTED VERSION, PUBLISHED AT \url{HTTPS://DOI.ORG/10.1109/TMRB.2026.3744947} — IEEE Transactions on Medical Robotics and Bionics}}%
{}

\IEEEpubid{}

\maketitle

\begin{abstract}
Free-hand dental procedures are typically repetitive, time-consuming and require high precision and manual dexterity.
Robots can play a key role in improving procedural accuracy and safety, enhancing patient comfort, and reducing operator workload.
However, robotic solutions for free-hand procedures remain limited or completely lacking. To address this gap, we develop a haptic bilateral teleoperation system (HBTS) for free-hand dental procedures (FH-HBTS).
The system includes a mechanical end-effector, compatible with standard clinical tools, and equipped with an endoscopic camera for improved visibility of the intervention site. By ensuring motion and force correspondence between the operator's and the robot's actions, monitored through visual feedback, we enhance the operator's sensory awareness and motor accuracy. Furthermore, to ensure procedural safety, we limit interaction forces by scaling the motion references provided to the admittance controller based solely on measured contact forces.
This ensures effective force limitation in all contact states without requiring prior knowledge of the environment. The proposed FH-HBTS is validated through a technical evaluation and an in-vitro pre-clinical study conducted on a dental model under clinically representative, yet simplified, conditions.
The results show that the system improves the naturalness, safety, and accuracy of teleoperation, highlighting its potential to enhance free-hand dental procedures.
\end{abstract}


\section{Introduction}\label{sec:introduction}
\IEEEPARstart{T}{he} World Health Organization estimates that oral diseases affect nearly 3.5 billion people worldwide.
However, access to primary oral health services is limited due to the high costs of procedures, the unequal distribution of oral health professionals, and the lack of adequate healthcare facilities \cite{grischke_dentronics_2020}.  

Traditionally, routine, prosthetic, periodontal, and some surgical dental procedures are performed free-hand. 
Routine procedures include the treatment of dental caries and endodontic therapy, while prosthetic procedures focus on the replacement of teeth through the use of prostheses.
These procedures involve the preparation of the tooth, which entails the removal of hard tissue using rotary instruments mounted on dental handpieces \cite{podhorsky_tooth_2015}. 
Periodontal procedures include scaling and root planing (SRP), i.e., the removal of plaque and calculus from the teeth and the smoothing of root surfaces, typically employing hand instruments and inserts mounted on sonic or ultrasonic handpieces \cite{sanz_treatment_2020}. 
Finally, although robotic computer-assisted implant surgery (r-CAIS) have shown a significant increase in the accuracy of dental implant placement, many operators still prefer the free-hand method for dental implantology \cite{afshari_free-hand_2022}.
Each of these procedures requires accuracy, extensively trained manual skills, and dexterity.
\begin{figure}[!t]
    \centering
    \includegraphics[width=0.9\columnwidth]{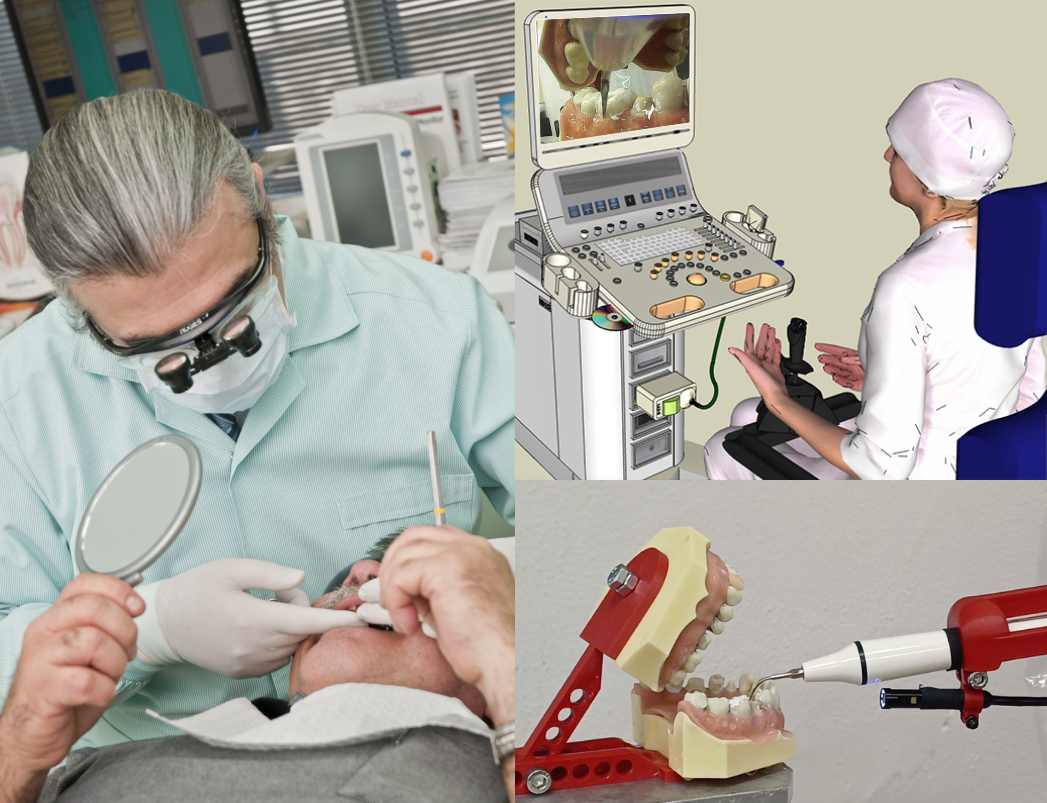}
    \caption{Manual (left) vs teleoperated (right) dental procedures (pictorial views).}
    \label{fig:dental-robots-advantages}
\end{figure}

\begin{remark}\label{rem:unstructured-environment}
    The oral cavity is a highly unstructured environment, which is full of natural obstacles that occlude the view of the surgical site and reduce the space for maneuvering, requiring continuous changes in the orientation of the clinical tool and the operator's posture.
\end{remark}
Therefore, the successful outcome of a procedure, as well as the comfort for the patient, depend on both human factors, i.e., the experience and skills of the human operator \cite{kozlovsky_influence_2018}, the adequacy of the instrumentation used and the duration of the treatments \cite{luo_patients_2018}, and environmental factors \cite{schnutenhaus_factors_2021}, as highlighted in Remark \ref{rem:unstructured-environment}. 
In particular, the latter factors can significantly affect the operator's comfort. In \cite{soo_occupational_2023} a high prevalence of musculoskeletal disorders among dentists is reported. 
The main causes of these disorders can be traced back to an incorrect application of ergonomics, such as awkward postures, repetitive and prolonged movements, use of small or narrow tool handles, exertion of forces on hard or sharp objects, and exposure to vibrations \cite{gupta_ergonomics_2014}. 
Hence, when training new operators, much emphasis is placed on learning skills related to space utilization and positioning of surgical and non-surgical instruments, e.g., mirrors, to overcome occlusions, while maintaining healthy and comfortable postures \cite{field_defining_2021}. 
In addition, to ensure the safety and effectiveness of the procedure, new operators must acquire skills related to the correct amount of pressure to apply to the site of interest, depending on the type of tool used.  
\begin{remark}\label{rem:sense-of-touch}
    The sense of touch plays a key role in achieving successful and efficient procedures.
\end{remark}
This is also apparent from emerging trends in teaching, which make use of virtual environments and haptic devices (HDs) for training new operators \cite{mcgleenon_preparing_2021, uoshima_technical_2021}.

In this context, the role of robots is to improve dental procedures, in terms of both patient satisfaction, i.e., broader access to care, reduced treatment time, and fast recovery, and operator comfort, i.e., reduced workload intensity, increased visibility of the surgical field, and improved ergonomics \cite{abouzeid_role_2021}, as shown in Figure \ref{fig:dental-robots-advantages}.

\subsection{Related works}\label{sec:related_works}
Haptic teleoperations are an enabling technology for the remote execution of complex tasks, employing a robotic system.
\begin{remark}\label{rem:haptic-teleoperations}
Haptic teleoperations combine the robot skills with the flexibility, dexterity, and cognitive capabilities of humans, while overcoming barriers in terms of distance, hazardous environments, and scales beyond human reach.
\end{remark}
By providing multiple types of sensory information, i.e.\ visual and haptic feedback, HBTSs enable the human operator to experience \textit{telepresence} \cite{raju_design_1989}.
Furthermore, by integrating different haptic feedback and guidance modalities \cite{yasin_evaluation_2021, kundrat_toward_2019}, and shared control approaches \cite{deng_shared_2024}, they contribute to reducing errors and enhance the efficiency, accuracy and safety of procedures, while also improving operator comfort and confidence.
For these reasons, such technology is consolidated in, e.g., telesurgery, space exploration, and the handling of hazardous materials in nuclear or underwater environments.
However, in the context of dentistry, its adoption is still very limited.

The majority of research in robotics for dentistry is focused on oral implantology \cite{alqutaibi_applications_2023}, mainly because robotic implant surgery offers significant benefits in terms of implant accuracy in partially \cite{wu_accuracy_2024} or totally edentulous patients.
Such technology also enables flapless implant placement, accelerating recovery time, reducing postoperative pain, and enhancing patient comfort \cite{talib_flapless_2022}. \cite{kasahara_telerobotic-assisted_2012} shows that a telerobotic-assisted drill can enhance accuracy while preserving awareness through an effective transmission of cutting forces and vibrations.

On the other hand, the state-of-the-art shows considerably less research effort in other free-hand procedures, a trend also highlighted in existing reviews of dental robotics \cite{ahmad_dental_2021}, which report a strong imbalance in favor of implantology. Despite this imbalance, the literature reports evidence of the potential benefits of robotic technologies in free-hand dental procedures. Such evidence is analyzed in the following and summarized in the supplementary materials\footnote{This paper has supplementary downloadable material available at http://ieeexplore.ieee.org, provided by the authors. This includes a supplementary PDF format information file and a multimedia MOV format movie clip, showing the end-to-end system experimental demonstration. This material is 74.4 MB in size.}. The analysis focuses on robotic systems explicitly developed for dental procedures and directly interacting with the oral environment, aiming at supporting or automating clinical free-hand procedures, while excluding works specifically addressing oral implantology. Only approaches supported by experimental, in vitro, or preliminary validation are considered, whereas purely virtual studies or robotic solutions not specifically tailored to dental procedures are not included.

In the context of prosthetic procedures, robotics can enhance the precision of manual grinding using a turbine drill during tooth preparation \cite{jiang_adjacent_2022} and can compensate for hand tremors and the lack of tactile feedback in laser ablation \cite{wang_preliminary_2013, wang_automatic_2014}. This can improve accuracy \cite{ma_trajectory_2013} and efficiency \cite{yuan_preliminary_2020}, and can enable automated tooth arrangement \cite{zhang_tooth_2001} and preparation for veneers and full crowns \cite{otani_vitro_2015, yuan_automatic_2016}.
In endodontics, robotics can automate and enhance precision and efficiency in root canal treatments \cite{dong_development_2009, dong_design_2012}, outperforming current approaches also in effectively addressing biological and mechanical traits associated with biofilm recalcitrance and reinfection \cite{hwang_catalytic_2019}, and can support visual-guided tool navigation \cite{gulrez_visual_2010}. 
In periodontal procedures, robots have the potential to increase accuracy, reducing the risk of errors and trauma to tissues by optimizing applied forces, as with other free-hand procedures. 
They can also improve the ergonomics for the operator, and comfort for the patient by reducing recovery time.
However, for such procedures, robotic technologies are still lacking \cite{liu_robotics_2023}, and there is no empirical evidence of such advantages.
Existing systems addressing force-regulated interaction with teeth are limited to adjacent tasks, such as tooth-brushing assistance \cite{ajani_hybrid_2020}, and are not designed for direct periodontal procedures.

Thus, although robotics shows promise in dental practice, several limitations remain. First, the quality of research is still insufficient in terms of clinical validation and technology readiness level (TRL) \cite{van_riet_robot_2021}. 
This is heavily influenced by the high costs of these systems and the need to ensure their compatibility with existing clinical tools \cite{xia_robotics_2024}. 
Second, the most widespread robots are autonomous, which require the development of advanced capabilities in perception, recognition, and exception handling \cite{xia_robotics_2024}. 
In this context, the presence of a human is still necessary not only as a supervisor but also as an operator.
In light of Remark \ref{rem:sense-of-touch} and Remark \ref{rem:haptic-teleoperations}, and considering the impact on the operator acceptance \cite{li_clinical_2023}, haptic teleoperations represent a viable solution. Moreover, HBTSs would allow the acquisition of human control strategies directly exercised in the robot state space, simplifying the process of transferring the knowledge from the human to the robot \cite{ravichandar_recent_2020} and enabling the effective and objective evaluation of operator skills during training \cite{li_optimized_2007}.
However, as emerges from the literature analysis above, no robotic solutions based on haptic bilateral teleoperation have been reported for free-hand tasks.
Finally, research tends to prioritize oral implantology over other dental procedures, despite the high number of daily free-hand routine procedures and the less invasiveness, entailing better patient acceptance \cite{milner_patient_2019}. Considering these limitations, \cite{ahmad_dental_2021} recommends to shift the focus from oral implantology towards less invasive procedures, which are better suited as forerunners in the field, and also to invest in educational robotics research as a valuable tool to overcome the barrier of acceptance of robotic systems among future dentists.

\subsection{Contributions}
Motivated by the limitations or complete absence of robotic solutions for the most repetitive free-hand dental procedures, and in light of the advantages introduced by haptic teleoperations, as in Remark \ref{rem:haptic-teleoperations}, we propose a HBTS for free-hand dental procedures (FH-HBTS). The proposed system features a custom designed end-effector, compatible with existing clinical tools, and a dedicated haptic control system that ensures procedural safety, improves accuracy, and preserves manual dexterity.
While building upon established teleoperation and interaction-control schemes, the proposed system extends existing HBTSs in two aspects where they remain limited.
Specifically, the key contributions are: (1) an eye-hand coordination controller providing both motion and force correspondence between the HD's workspace and the video stream coming from the eye-in-hand camera; with respect to existing HBTSs, such mapping is bilateral and drift-free, even after prolonged teleoperation sessions; (2) a force limitation strategy effectively reducing interaction forces; with respect to existing HBTSs, this strategy needs no previous assumption or online estimation of the environment geometry, therefore accounting for the challenges outlined in Remark \ref{rem:unstructured-environment}; notably, this strategy is part of the onboard robot controller and is therefore compatible with passivity-based teleoperations; (3) an integrated TRL-5 system for free-hand dental procedures, validated on SRP, demonstrating technical feasibility and clinical suitability; with respect to existing HBTSs, no prior system has been integrated and validated specifically for free-hand dental procedures.

In light of the inclusion and exclusion criteria defined in Section \ref{sec:related_works}, and also reported in the supplementary materials, to the best of the authors' knowledge, this is the first HBTS specifically complying with the special requirements of safety, accuracy, and dexterity characterizing free-hand dental procedures.

\section{Preliminaries}\label{sec:preliminaries}
\subsection{Notation}\label{sec:notation}
We denote by ${\bf T}_{b}^{a} \in SE(3)$ the homogeneous transformation matrix from frame $a$ to frame $b$, in the form:
\begin{equation} \label{eq:transform-matrix}
    {\bf T}_{b}^{a} =
    \begin{pmatrix}
        {\bf R}_{b}^{a} & {\bf p}_{b}^{a} \\
        {\bf 0}^\top & 1
    \end{pmatrix},
\end{equation}
where ${\bf R}_{b}^{a} \in SO(3)$ is the rotation matrix from frame $a$ to frame $b$, with ${\bf R}_{a}^{b} = ({\bf R}_{b}^{a})^{-1} \equiv ({\bf R}^{a}_{b})^\top$, and ${\bf p}_{b}^{a} \in \mathbb{R}^{3 \times 1}$ represents $b$'s origin in frame $a$. 

We let ${\bf x}_{b}^{a} = \begin{bmatrix}({\bf p}_{b}^{a})^\top, ({\bf q}_{b}^{a})^\top \end{bmatrix}^\top$ be the pose of the frame $b$ with respect to the frame $a$, corresponding to ${\bf T}_{b}^{a}$, where ${\bf q}_{b}^{a} \in \mathbb H_1$ is the quaternion associated with ${\bf R}_{b}^{a}$.

We denote by ${\bf R}_{\bf a}^{\mathcal F}(\alpha) \in SO(3)$ the elementary rotation of an angle $\alpha$ about a generic axis $\bf a$ of a frame $\mathcal F$. 

We denote by ${\bf v}^{a}_{b} = \begin{bmatrix}(\dot{\bf p}^{a}_{b})^\top, ({\boldsymbol \omega}^{a}_{b})^\top \end{bmatrix}^\top \in \mathbb R^{6 \times 1}$ the twist of the frame $b$ with respect to the frame $a$, where $\dot{\bf p}^{a}_{b}, {\boldsymbol \omega}^{a}_{b} \in \mathbb{R}^{3 \times 1}$ are the linear and angular velocities, respectively.

We let ${\bf h}^{a} = \begin{bmatrix}({\bf f}^{a})^\top, ({\bf m}^{a})^\top \end{bmatrix}^\top \in \mathbb R^{6 \times 1}$ be the wrench with respect to the frame $a$, where ${\bf f}^{a}, {\bf m}^{a} \in \mathbb{R}^{3 \times 1}$ are the forces and momenta, respectively.

${\bf I}_n$ indicates the $n \times n$ identity matrix.

\subsection{Haptic device and robot reference frames} \label{haptic-device-and-follower-robot-reference-frames}
\begin{figure}
    \centering
    \subfloat{\includegraphics[width=0.40\columnwidth]{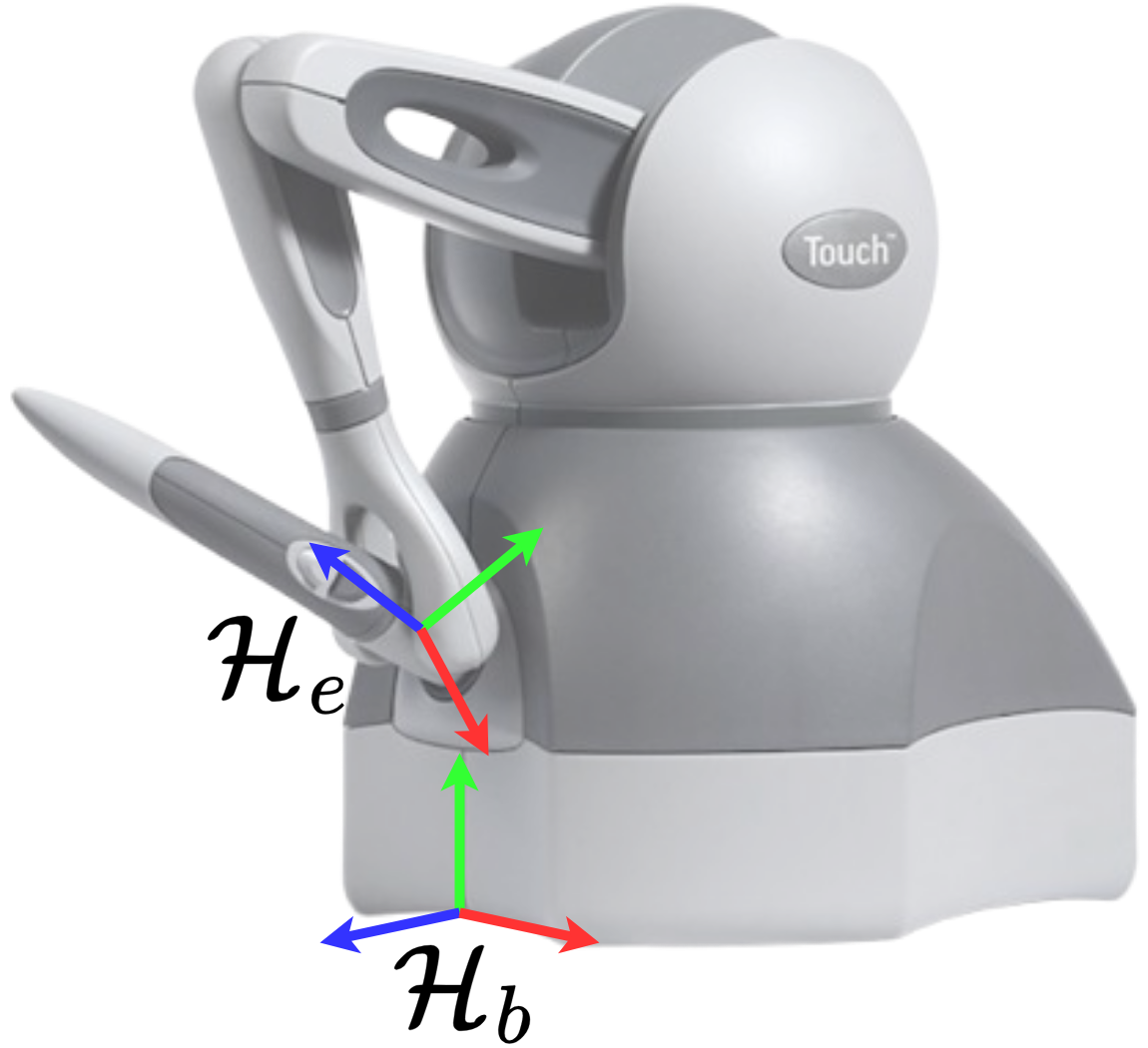}}
    \hfill
     \subfloat{\includegraphics[width=0.40\columnwidth]{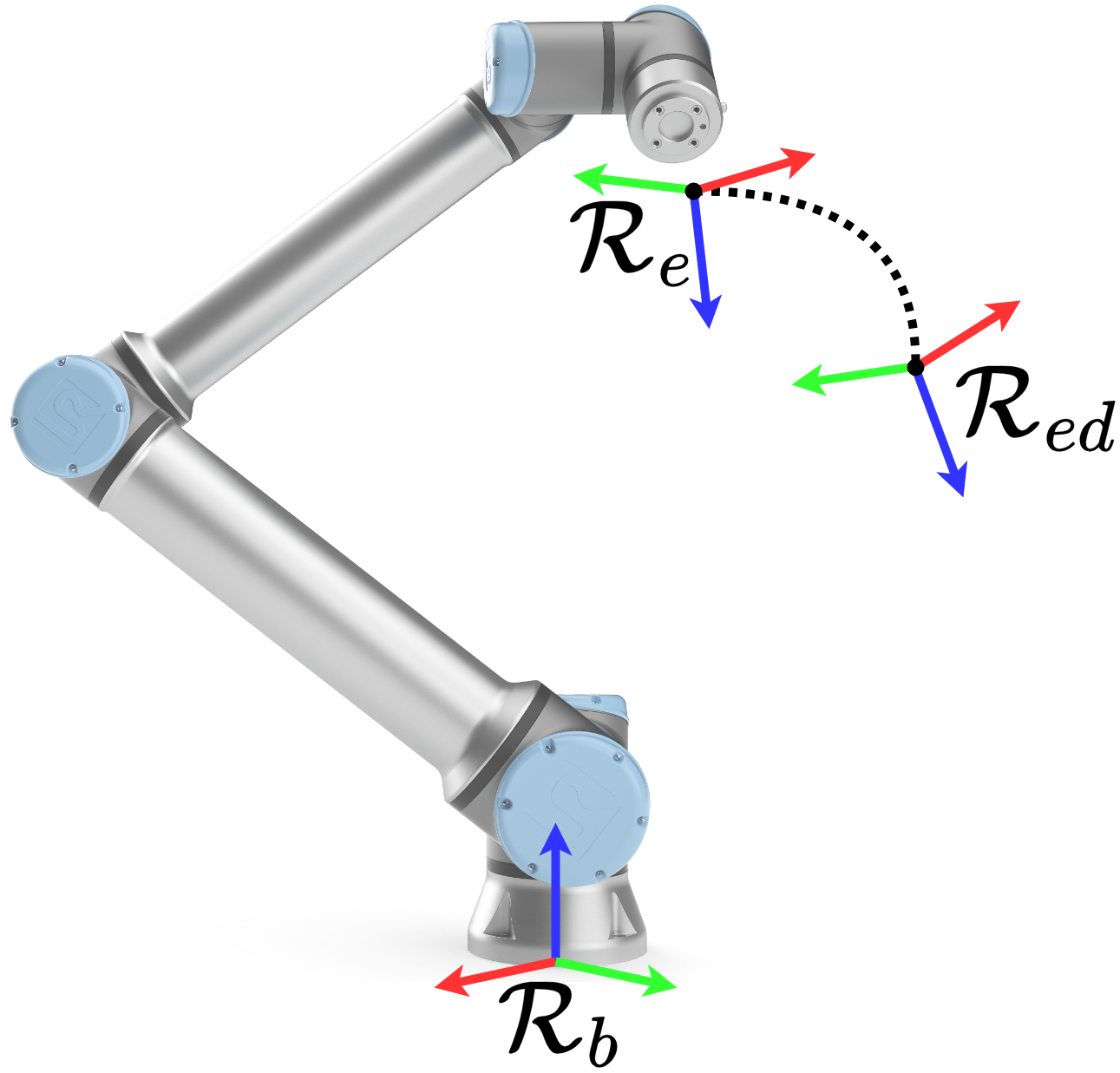}}
    \caption{Haptic device and robot reference frames.}
    \label{fig:hd-robot-frames}
\end{figure}

The frames used by the HD and the robot are reported in Figure \ref{fig:hd-robot-frames}. With regard to the HD, two frames are introduced: $\mathcal H_b$ is the base reference frame, in which the HD's stylus poses and the feedback forces are expressed; $\mathcal H_e$ is the HD's stylus frame.
Regarding the robot, four frames are defined: $\mathcal R_b$ is the base reference frame, in which motion references are expressed; $\mathcal R_e$ is the robot's end-effector frame (computed from forward kinematics of joint measurements), corresponding to the tool center point (TCP); $\mathcal R_{ed}$ is the desired robot's end-effector frame, input to admittance control; $\mathcal R_c$ is the compliant frame, output by the admittance control law. 

We note that, after admittance control, motion control guarantees that $\mathcal R_e$ tracks $\mathcal R_c$, as discussed next.

\section{Methodology}\label{sec:methodology}
\begin{figure*}
    \centering
    \includegraphics[width=\textwidth]{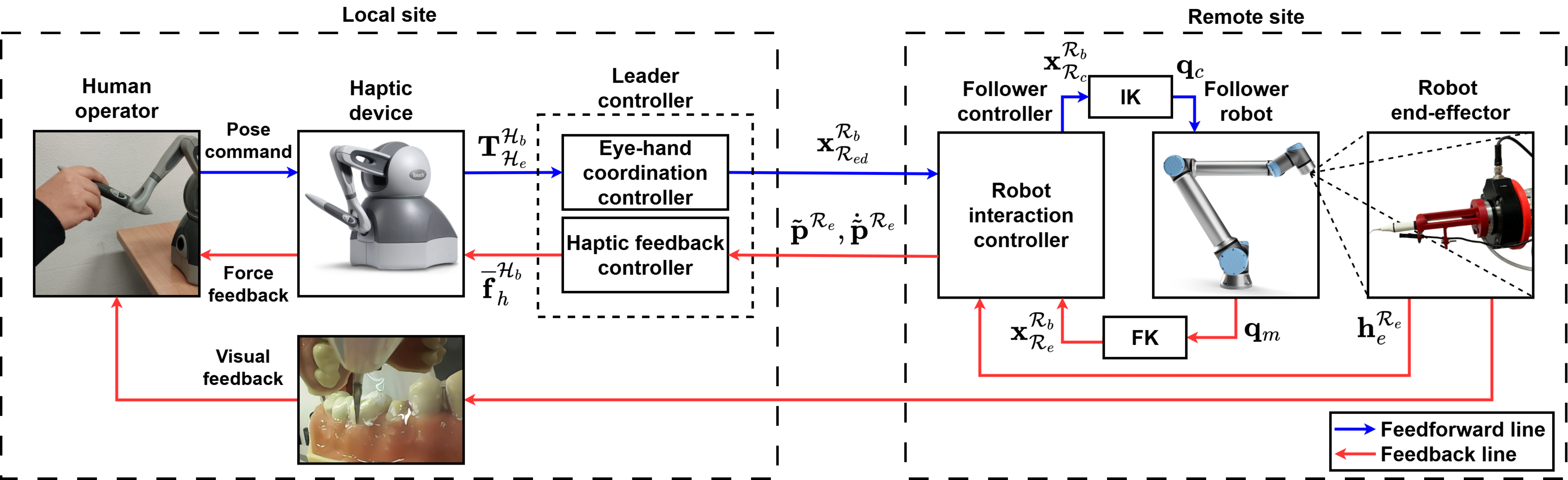}
    \caption{Haptic bilateral teleoperation system scheme.}
    \label{fig:HBTS}
\end{figure*}

The FH-HBTS we propose, shown in Figure \ref{fig:HBTS}, is based on a position-position architecture \cite{niemeyer_telerobotics_2016}, coupling the leader (LR) and follower (FR) robots by means of a spring–damper, which is a passive element.
Therefore, in the absence of communication delays, this architecture has the advantage of being intrinsically stable \cite{aliaga_experimental_2004, hashtrudi-zaad_analysis_2001}. A formal proof of absolute stability can be found in \cite{razi_extension_2012}.
In the proposed design, we assume negligible time delays and packet losses and, accordingly, do not integrate passivity-based control approaches \cite{panzirsch_exploring_2022}.
This choice is motivated both by the validity of this assumption in healthcare facilities and by the conservative nature of passivity-based methods, which are known to worsen transparency and system usability in order to ensure stability under adverse network conditions \cite{salehi_improvement_2024}.
Indeed, in precision free-hand dental procedures, this trade-off is generally undesirable, as it may impair fine force discrimination and the naturalness of the interaction.
Nevertheless, the management of communication delays and packet losses represents an orthogonal direction to the one investigated in this work. Indeed, under non-ideal network conditions, the proposed architecture can be readily extended to enforce time-domain passivity of the communication channel, for instance by adopting a TDPA-based approach \cite{artigas_time_2010}, as demonstrated in Section \ref{sec:force-limitation-assessment}.

The LR is an impedance HD, sensing the pose of the operator and rendering haptic feedback according to the remote interaction between the FR and the environment. The HD is controlled by a leader controller, featuring an eye-hand coordination controller (EHCC) and a haptic feedback controller (HFC). The former is responsible for capturing the HD's stylus poses, filtering human tremors, and generating the corresponding robot end-effector poses, so that the robot follows the stylus motion in coordination with the visual feedback. The latter is responsible for rendering the force feedback on the HD to let the human operator perceive the remote environment. Both controllers record data throughout the entire procedure. 
The FR is a position-controlled robot, instructed to track the HD through an admittance controller, which provides a configurable degree of compliance on its target pose with respect to the environment and improves safety through dedicated mechanisms and emergency protocols.

With reference to Figure \ref{fig:HBTS}, depicting the complete system, the following sections examine its individual subcomponents in detail.
Section \ref{sec:eye-hand-coordination-controller} and Section \ref{sec:haptic-feedback-controller} provide a detailed description of the objectives and operation of the EHCC and HFC controllers.
Section \ref{sec:robot-interaction-control} describes the robot interaction control.
Finally, Section \ref{sec:integrated-mechanical-end-effector} presents the design of the mechanical end-effector for the FR.

\subsection{Eye-hand coordination controller}\label{sec:eye-hand-coordination-controller}
In a HBTS, visual feedback plays a crucial role in overcoming the limitations caused by distance. 
In teleoperated free-hand dental procedures, using an eye-in-hand camera is the most effective solution \cite{flandin_eye--handeye--hand_2000}. 
This approach enables the operator to maintain a consistently centered view of the operative field, offering a significant advantage over eye-to-hand camera configurations, where the view may be off-center with respect to the area of interest or obstructed by physical obstacles. 
In this context, the human operator relies on the video of the camera to command the FR via the HD. 
However, in the eye-in-hand configuration, a rotation of the tool (unavoidable, to preserve dexterity) causes a rotation of the viewing angle, whose lack of compensation results in a loss of both motion and force correspondence between the movements and perceived interaction expected by the operator and those actually executed by the robot in the camera view.
This highlights the importance of an appropriate coupling.

The literature shows that this coupling can be achieved through simple kinematic transformations and scaling strategies \cite{radi_workspace_2012}.
Common mapping strategies include position control, workspace drift control \cite{conti_spanning_2005}, and rate control \cite{salcudean_transparent_2000} which define how the operator's motions are translated into robot commands, complemented by force scaling \cite{radi_workspace_2012} to map the interaction forces into the range of forces that the HD can exert.
However, these methods generally neglect viewing angle effects, relying on display-level compensation and thus compromising force correspondence.
Indeed, the da Vinci Surgical System\textsuperscript{TM}, which requires the insertion of multiple tools to implement an eye-to-hand strategy, thereby increasing bulk and cost of the equipment, provides the motion correspondence by projecting the endoscope image above the surgeon's hands by means of mirrored overlay optics \cite{freschi_technical_2013}. 
In a similar manner, in eye-in-hand cases, motion correspondence is typically achieved by aligning the visual feedback with the operator's control frame through online compensation of camera–control frame misalignment in the displayed view \cite{wu_camera_2023}.

To overcome this limitation, we introduce an eye–hand coordination mapping preserving both motion and force correspondence. 
Within the proposed EHCC, this mapping is implemented through a view-consistent transfer of the HD twist to the FR and incorporates dynamic scaling factors to improve control accuracy according to task requirements.
To enhance the naturalness of the teleoperation, a rotational coupling is also established between the HD's stylus and the FR's TCP. This includes defining the desired relative rotation ${\bf R}_{\mathcal R_e}^{\mathcal H_e}$, and aligning manually.
This is done by the human operator during what we indicate as the \textit{engagement phase}. 

The latter is presented in Section \ref{sec:engagement-phase}, while the human tremors filtering and the eye-hand coordination mapping, implementing the motion correspondence, are discussed in Section \ref{sec:human-tremors-filtering} and Section \ref{sec:eye-hand-coordination}, respectively.
The eye-hand coordination implementing the force correspondence is described in Section \ref{sec:haptic-feedback-controller}.

\subsubsection{Engagement phase}\label{sec:engagement-phase}
\begin{figure}
    \centering
    \subfloat{\includegraphics[width=0.45\columnwidth]{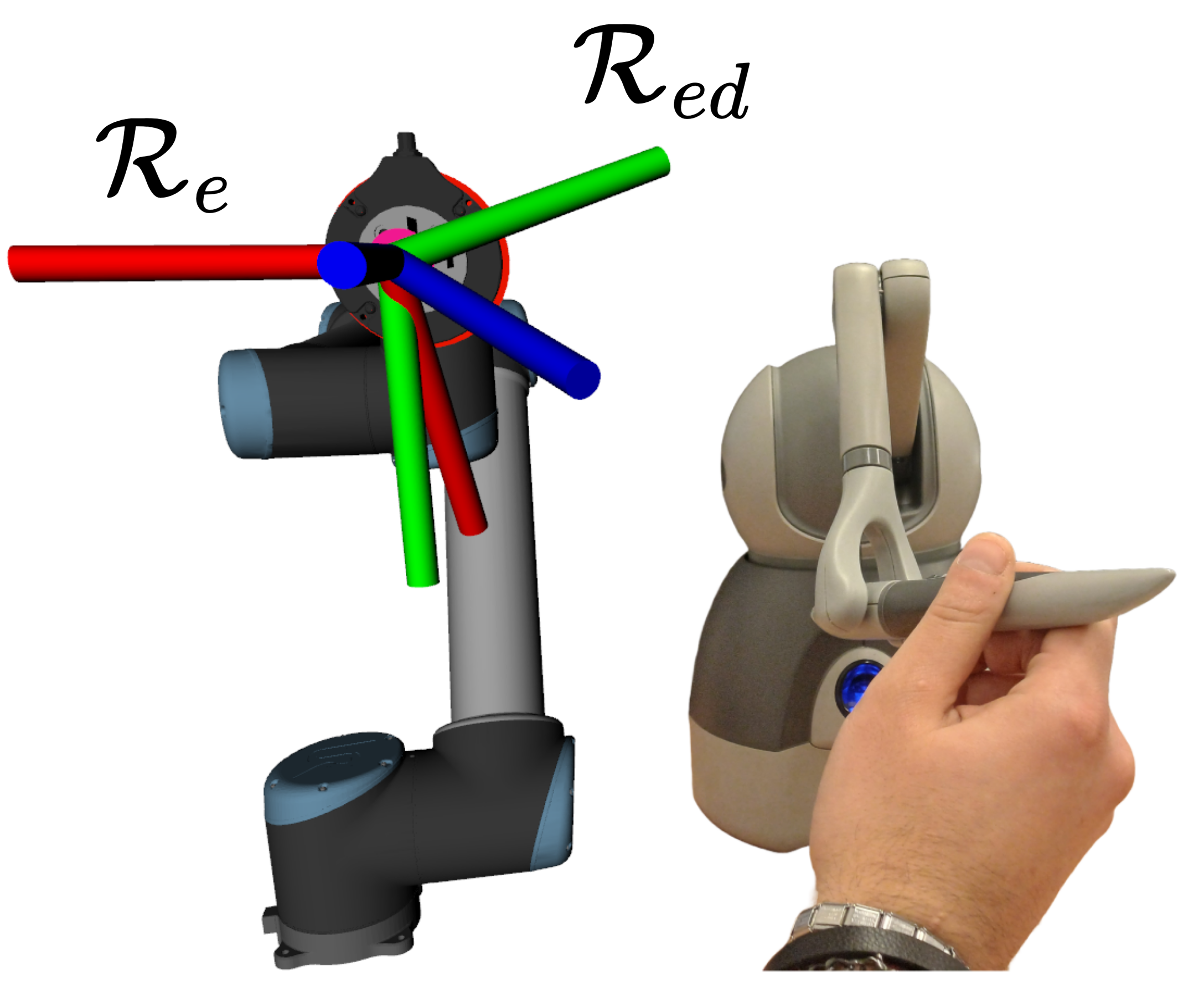}}
    \hfill
     \subfloat{\includegraphics[width=0.42\columnwidth]{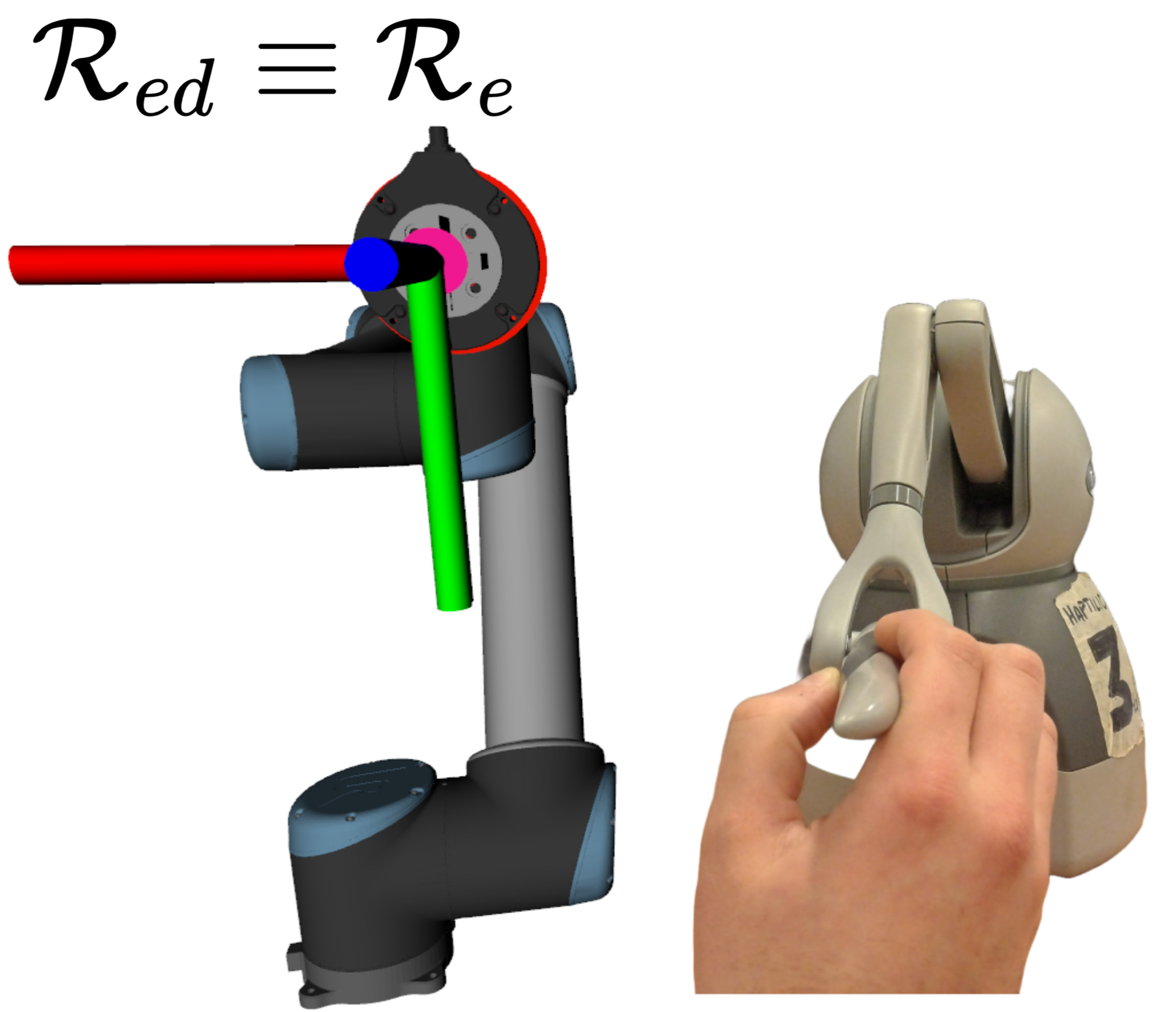}}
    \caption{Engagement phase supported by the visualization tool.}
    \label{fig:engagement-phase}
\end{figure}

During the engagement phase, the EHCC retrieves the homogeneous transformation matrix of the HD's stylus ${\bf T}_{\mathcal H_e}^{\mathcal H_b}(k)$, and converts it into a robot's end-effector homogeneous transformation matrix with the following kinematic equations:
\begin{align}
    {\bf R}_{\mathcal R_{ed}}^{\mathcal R_b}(k) &= {\bf R}_{\mathcal H_b}^{\mathcal R_b} {\bf R}_{\mathcal H_e}^{\mathcal H_b}(k) {\bf R}_{\mathcal R_e}^{\mathcal H_e},\label{eq:HD-converted-rotation}\\
   {\bf p}_{\mathcal R_{ed}}^{\mathcal R_b}(k) &= {\bf p}_{\mathcal R_e}^{\mathcal R_b}(k),\label{eq:HD-converted-position}
\end{align}
where ${\bf R}_{\mathcal H_b}^{\mathcal R_b}$ is the fixed rotation between $\mathcal H_b$ and $\mathcal R_b$.
We note that the position of the HD's stylus is not relevant for coupling, so it is converted to the current position of the robot's TCP.
The obtained frame $\mathcal R_{ed}$ is real-time displayed in a visualization tool, together with the actual robot end-effector one $\mathcal R_e$, as shown in Figure \ref{fig:engagement-phase}, so that the human operator, by acting on $\mathcal{H}_e$, can align the frame $\mathcal R_{ed}$ with the frame $\mathcal R_e$. Once the alignment is accomplished, i.e., ${\bf R}_{\mathcal R_b}^{\mathcal R_e}(k){\bf R}_{\mathcal R_{ed}}^{\mathcal R_b}(k) = {\bf I}_3$, the engagement phase ends, i.e., HD and FR are aligned, and the human operator can begin teleoperation.  

\subsubsection{Human tremors filtering}\label{sec:human-tremors-filtering}
At each control cycle, the EHCC retrieves the actual pose ${\bf x}_{\mathcal H_e}^{\mathcal H_b}(k)$ of the HD's stylus, corresponding to ${\bf T}_{\mathcal H_e}^{\mathcal H_b}(k)$. Exploiting such a pose to generate a command for the robot would result in transferring the human hand tremors, further accentuated by the control frequency. Therefore, ${\bf x}_{\mathcal H_e}^{\mathcal H_b}(k)$ is filtered online by a moving average filter for poses. At the $k$-th control cycle, the filtered position is obtained as $\overline{{\bf p}}_{\mathcal H_e}^{\mathcal H_b}(k) = {\bf s}(k) / m$, where $m \leq n$ is the number of collected samples at the $k$-th control cycle in the moving average window of size $n$, and ${\bf s}(k)$ is their cumulative sum:
\begin{equation}\label{eq:moving-average-filter}
    {\bf s}(k) = {\bf s}(k-1) + {\bf p}_{\mathcal H_e}^{\mathcal H_b}(k) - \begin{cases}
        0, & \text{if}~k < n; \\
        {\bf p}_{\mathcal H_e}^{\mathcal H_b}(k - n), & \text{otherwise.}
    \end{cases}
\end{equation}

The filtered quaternion $\overline{\bf q}_{\mathcal H_e}^{\mathcal H_b}(k)$ is the eigenvector associated with the maximum eigenvalue of ${\bf M}(k) = {\bf C}(k) / m$ \cite{markley_averaging_2007}, where ${\bf C}(k)$ is the cumulative sum of samples within the window, updated at each control cycle:
\begin{equation}
    {\bf C}(k) = \begin{cases}
        {\bf C}(k-1) + {\bf q}_{\mathcal H_e}^{\mathcal H_b}(k)\bigl[{\bf q}_{\mathcal H_e}^{\mathcal H_b}(k)\bigr]^\top, & \text{if}~k < n; \\
        {\bf C}(k-1) + {\bf q}_{\mathcal H_e}^{\mathcal H_b}(k)\bigl[{\bf q}_{\mathcal H_e}^{\mathcal H_b}(k)\bigr]^\top + & \\
        \quad - {\bf q}_{\mathcal H_e}^{\mathcal H_b}(k - n)\bigl[{\bf q}_{\mathcal H_e}^{\mathcal H_b}(k - n)\bigr]^\top, & \text{otherwise.}
    \end{cases}
\end{equation}
Keeping the cumulative sums ${\bf s}(k)$ and ${\bf C}(k)$ enables achieving $\mathcal{O}(1)$ computational complexity. 

\subsubsection{Eye-hand coordination}\label{sec:eye-hand-coordination}
Let $\Delta \overline{{\bf p}}_{\mathcal H_e}^{\mathcal H_b}(k)$ be the HD's displacement:
\begin{equation}
    \Delta \overline{{\bf p}}_{\mathcal H_e}^{\mathcal H_b}(k) = \overline{{\bf p}}_{\mathcal H_e}^{\mathcal H_b}(k) - \overline{{\bf p}}_{\mathcal H_e}^{\mathcal H_b}(k-1),
\end{equation}
and $\overline{{\bf R}}^{\mathcal H_e(k-1)}_{\mathcal H_e(k)}$ the relative rotation
obtained from
\begin{equation}
    \overline{\bf q}_{\mathcal H_e(k)}^{\mathcal H_e(k-1)} = \bigl[\overline{\bf q}_{\mathcal H_e}^{\mathcal H_b}(k-1)\bigr]^\top \overline{\bf q}_{\mathcal H_e}^{\mathcal H_b}(k).
\end{equation}
At each control cycle, the EHCC computes the HD's stylus twist:
\begin{equation}\label{eq:HD-twist}
    {\bf v}_{\mathcal H_e}^{\mathcal H_b}(k) = T_s^{-1} \begin{bmatrix}\Delta \overline{{\bf p}}_{\mathcal H_e}^{\mathcal H_b}(k)^\top, & \vartheta{\bf r}^\top\end{bmatrix}^\top,
\end{equation}
where $T_s \in \mathbb R_+$ is the control period, and ${\bf r} \in \mathbb R^{3 \times 1}$ and $\vartheta \in \mathbb R$ are the axis-angle representation corresponding to $\overline{{\bf R}}^{\mathcal H_e(k-1)}_{\mathcal H_e(k)}$, respectively.

\begin{figure}
    \centering
    \subfloat{\includegraphics[width=0.47\columnwidth]{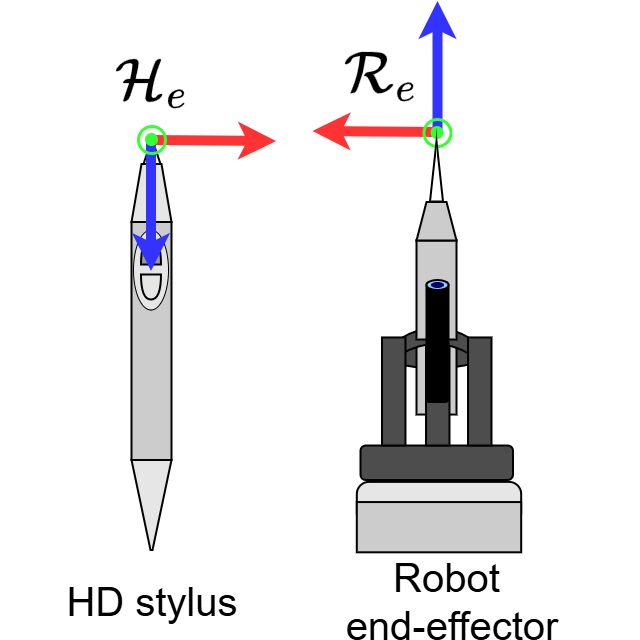}}
     \subfloat{\includegraphics[width=0.47\columnwidth]{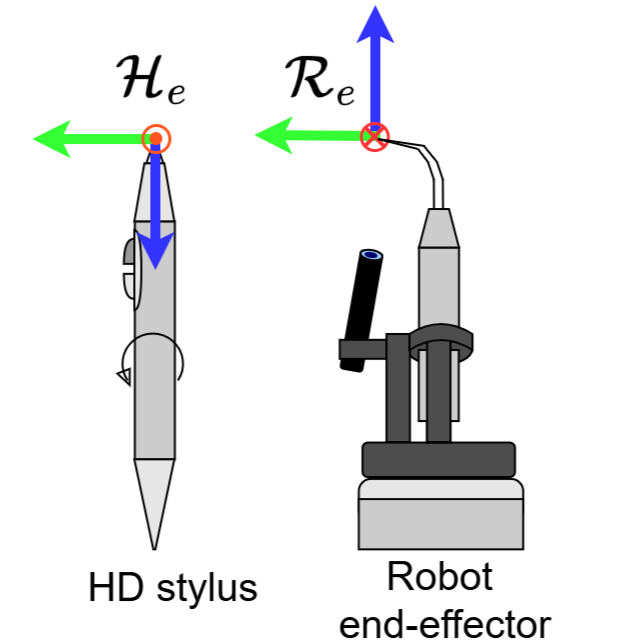}}
     \\
     \subfloat{\includegraphics[width=0.43\columnwidth]{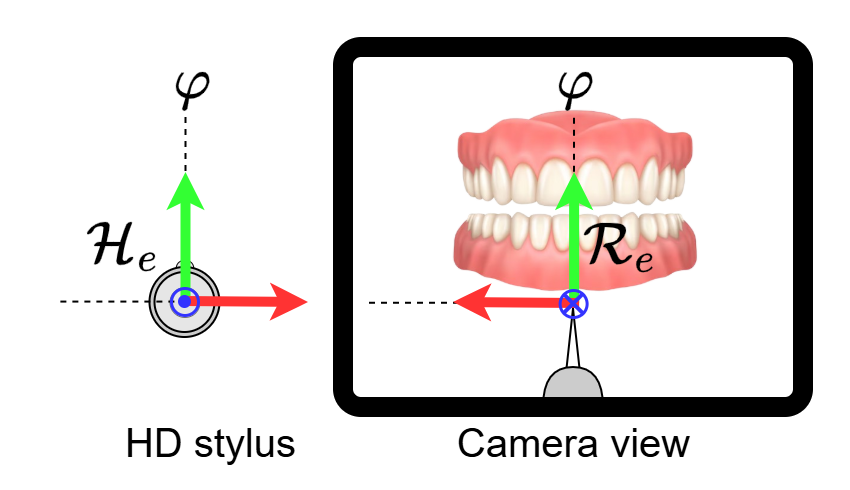}}
     \subfloat{\includegraphics[width=0.43\columnwidth]{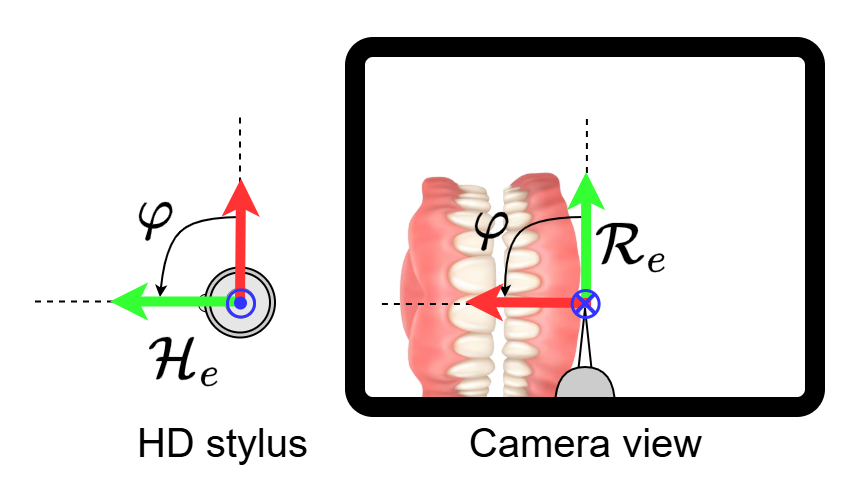}}
    \caption{Visualization of the rotational coupling between $\mathcal H_e$ and $\mathcal R_e$: (top) top view; (bottom) user view; (left) $\varphi=0$; (right) $\varphi=\pi/2$.}
    \label{fig:rotational-mapping-visualization}
\end{figure}

From \eqref{eq:HD-twist}, the corresponding desired FR's end-effector twist is obtained as:
\begin{equation}\label{eq:FR-twist}
    \begin{split}
        {\bf v}_{\mathcal R_{ed}}^{\mathcal R_b}(k) = &\bigl({\bf I}_2 \otimes {\bf R}_{\mathcal R_e}^{\mathcal R_b}(k)\bigr) \biggl\{{\boldsymbol \Lambda} \Bigl[\bigl({\bf I}_2 \otimes {\bf R}_{\mathcal H_e}^{\mathcal R_e}(k)\bigr) \\
        \cdot & \bigl({\bf I}_2 \otimes {\bf R}_{\mathcal H_b}^{\mathcal H_e}(k)\bigr)
        {\bf v}_{\mathcal H_e}^{\mathcal H_b}(k)\Bigr]\biggr\}.
    \end{split}
\end{equation}
where $\otimes$ denotes the Kronecker product, ${\boldsymbol \Lambda} \in \mathbb{R}^{6 \times 6}$ is the scaling diagonal matrix with ${\boldsymbol \Lambda}_{4:6, 4:6} = {\bf I}_3$, and
\begin{equation}\label{eq:sr-space-to-hd-space}
    {\bf R}_{\mathcal H_e}^{\mathcal R_e}(k) = {\bf R}_{\bf z}^{\mathcal R_e}\bigl(\varphi(k)\bigr) {\bf R}_{\mathcal H_e}^{\mathcal R_e}.
\end{equation}
We note that ${\bf R}_{\mathcal{H}_e}^{\mathcal{R}_e}$ is the fixed rotation defining the mapping between quantities in $\mathcal{H}_e$ and $\mathcal{R}_e$, and $\bf z$ is the axis of $\mathcal R_e$, which we assume to coincide with the optical axis of the camera ${\bf o}$ in $\mathcal{R}_e$ for negligible values of the camera tilt angle. 
Therefore, a rotation of $\varphi(k)$ about the ${\bf z}$ axis of $\mathcal{R}_e$ results in a viewing angle of $\varphi(k)$.
The $\bf z$ axis of $\mathcal R_e$ is mapped to the ${\bf z}$ axis of $\mathcal{H}_e$, according to ${\bf R}_{\mathcal{H}_e}^{\mathcal{R}_e}$.
Hence, in the absence of compliance along rotational axes, the same behavior is achieved by rotating of $\varphi(k)$ about the ${\bf z}$ axis of $\mathcal{H}_e$, as shown in Figure \ref{fig:rotational-mapping-visualization}, and \eqref{eq:sr-space-to-hd-space} can be equivalently written as:
\begin{equation}\label{eq:sr-space-to-hd-space-1}
    {\bf R}_{\mathcal H_e}^{\mathcal R_e}(k) = {\bf R}_{\mathcal H_e}^{\mathcal R_e} {\bf R}_{\bf z}^{\mathcal H_e}\bigl(\varphi(k)\bigr).
\end{equation}

The rotation ${\bf R}_{\bf z}^{\mathcal R_e}\bigl(\varphi(k)\bigr)$, or equivalently ${\bf R}_{\bf z}^{\mathcal H_e}\bigl(\varphi(k)\bigr)$, ensures the motion correspondence between the operator's movements and those of the FR's TCP in the camera view, by appropriately rotating the desired velocities according to the viewing angle. By setting ${\bf R}_{\bf z}^{\mathcal R_e}(\varphi(k)) = {\bf I}_3$ (or equivalently ${\bf R}_{\bf z}^{\mathcal H_e}(\varphi(k)) = {\bf I}_3$) the mapping between twists becomes viewing angle independent:
\begin{equation}
    {\bf R}_{\mathcal H_e}^{\mathcal R_e}(k) = {\bf R}_{\mathcal H_e}^{\mathcal R_e}, \; \forall k \in \mathbb N_0.
\end{equation}
We note that the angle $\varphi(k)$ is obtained from the exact measurements provided by the FR's forward kinematics (or equivalently by the HD's).
For this reason, even during prolonged procedures, no orientation drift is accumulated, and motion correspondence is consistently preserved.

The desired FR's TCP homogeneous transformation ${\bf T}_{\mathcal R_{ed}}^{\mathcal R_b}(k)$ is obtained by numerically integrating the desired twist in \eqref{eq:FR-twist}. As for the desired rotation ${\bf R}_{\mathcal R_{ed}}^{\mathcal R_b}(k)$, it is
\begin{equation}\label{eq:angular-velocity-integration}
    {\bf R}_{\mathcal R_{ed}}^{\mathcal R_b}(k) = {\bf R}_{\mathcal R_{e}}^{\mathcal R_b}(k-1) {\bf R}_{\mathcal R_{ed}(k)}^{\mathcal R_e(k-1)},
\end{equation}
where ${\bf R}_{\mathcal R_{ed}(k)}^{\mathcal R_e(k-1)}$ is the rotation matrix corresponding to the axis $\frac{{\bf r}}{\lVert {\bf r} \rVert}$ and angle $\vartheta$:
\begin{equation}
    {\bf r} = {\bf R}_{\mathcal R_{b}}^{\mathcal R_e}(k-1) \bigl( {\boldsymbol \omega}_{\mathcal R_{ed}}^{\mathcal R_b}(k) T_s \bigr), \quad
    \vartheta = \left\lVert  {\bf r} \right\rVert.
\end{equation}
As for the desired position ${\bf p}_{\mathcal R_{ed}}^{\mathcal R_b}(k)$, it is obtained as:
\begin{equation}\label{eq:linear-velocity-integration}
    {\bf p}_{\mathcal R_{ed}}^{\mathcal R_b}(k) = {\bf p}_{\mathcal R_{ed}}^{\mathcal R_b}(k-1) + \dot{\bf p}_{\mathcal R_{ed}}^{\mathcal R_b}(k) T_s.
\end{equation}
From ${\bf T}_{\mathcal R_{ed}}^{\mathcal R_b}(k)$ the desired FR's TCP pose is obtained as:
\begin{equation}
    {\bf x}_{\mathcal R_{ed}}^{\mathcal R_b}(k) = \begin{bmatrix}
        {\bf p}_{\mathcal R_{ed}}^{\mathcal R_b}(k)^\top, & {\bf q}_{\mathcal R_{ed}}^{\mathcal R_b}(k)^\top
    \end{bmatrix}^\top.
\end{equation}
The eye-hand coordination mapping and scaling can be turned into an algorithmic procedure as described in Algorithm \ref{alg:ehc-mapping-and-scaling}. 
\begin{algorithm}
\caption{Eye-hand coordination mapping and scaling}\label{alg:ehc-mapping-and-scaling}
\begin{algorithmic}[1]
    \Require The HD's actual $\overline{\bf x}_{\mathcal H_e}^{\mathcal H_b}(k)$ and previous $\overline{\bf x}_{\mathcal H_e}^{\mathcal H_b}(k-1)$ filtered poses, and the previous desired pose ${\bf x}_{\mathcal R_{ed}}^{\mathcal R_b}(k-1)$
    \Ensure The desired FR's TCP pose ${\bf x}_{\mathcal R_{ed}}^{\mathcal R_b}(k)$
    \State Compute ${\bf v}_{\mathcal H_e}^{\mathcal H_b}(k)$ using \eqref{eq:HD-twist}
    \State Compute ${\bf v}_{\mathcal R_{ed}}^{\mathcal R_b}(k)$ using \eqref{eq:FR-twist}:
    \begin{enumerate}
        \item Convert ${\bf v}_{\mathcal H_e}^{\mathcal H_b}(k)$ into ${\bf v}_{\mathcal R_{ed}}^{\mathcal R_e}(k)$ and rotate it according to the viewing angle using \eqref{eq:sr-space-to-hd-space}:
        \begin{equation*}
            {\bf v}_{\mathcal R_{ed}}^{\mathcal R_e}(k):= \bigl({\bf I}_2 \otimes {\bf R}_{\mathcal H_e}^{\mathcal R_e}(k)\bigr) \bigl({\bf I}_2 \otimes {\bf R}_{\mathcal H_b}^{\mathcal H_e}(k)\bigr) {\bf v}_{\mathcal H_e}^{\mathcal H_b}(k)
        \end{equation*}
        \item Scale ${\bf v}_{\mathcal R_{ed}}^{\mathcal R_e}(k)$ using $\boldsymbol \Lambda$ and convert it into ${\bf v}_{\mathcal R_{ed}}^{\mathcal R_b}(k)$:
        \begin{equation*}
            {\bf v}_{\mathcal R_{ed}}^{\mathcal R_b}(k) = \bigl({\bf I}_2 \otimes {\bf R}_{\mathcal R_e}^{\mathcal R_b}(k)\bigr) \bigl({\boldsymbol \Lambda} {\bf v}_{\mathcal R_{ed}}^{\mathcal R_e}(k)\bigr)
        \end{equation*}
    \end{enumerate}
    \State Compute ${\bf T}_{\mathcal R_{ed}}^{\mathcal R_b}(k)$ from ${\bf v}_{\mathcal R_{ed}}^{\mathcal R_b}(k)$ using \eqref{eq:angular-velocity-integration} and \eqref{eq:linear-velocity-integration}
    \State Extract ${\bf x}_{\mathcal R_{ed}}^{\mathcal R_b}(k)$ from ${\bf T}_{\mathcal R_{ed}}^{\mathcal R_b}(k)$
\end{algorithmic}
\end{algorithm}

\subsection{Robot interaction control}\label{sec:robot-interaction-control}
In this section, we first formalize the contact model between the robot and the environment (Section \ref{sec:admittance-model}), and subsequently, we present the free-hand strategy for limiting the interaction forces (Section \ref{sec:free-hand-force-limitation-strategy}) and the other safety mechanisms embedded in the controller (Section \ref{sec:safety-mechanisms}).

\subsubsection{Admittance model}\label{sec:admittance-model}
Let ${\bf x}_{\mathcal R_{ed}}^{\mathcal R_b}(k)$ be the desired pose and $\mathcal R_c$ the compliant frame specified by the quantities ${\bf p}_{\mathcal R_c}^{\mathcal R_{ed}}(k)$, ${\bf q}_{\mathcal R_c}^{\mathcal R_{ed}}(k)$, ${\bf v}_{\mathcal R_c}^{\mathcal R_{ed}}(k)$, $\dot{\bf v}_{\mathcal R_c}^{\mathcal R_{ed}}(k)$.
When the FR interacts with the environment, $\mathcal R_c$ can be completely defined in $\mathcal R_{ed}$ by integrating the mechanical impedance equation:
\begin{equation}\label{eq:admittance-model}
    {\bf M}_d \dot{{\bf v}}^{\mathcal R_{ed}}_{\mathcal R_c}(k) + {\bf K}_D {{\bf v}}^{\mathcal R_{ed}}_{\mathcal R_c}(k) + {\bf K}_P {\bf x}^{\mathcal R_{ed}}_{\mathcal R_c}(k) = - \tilde{\bf h}^{\mathcal R_{ed}}(k),
\end{equation}
where ${\bf M}_d, {\bf K}_D, {\bf K}_P \in \mathbb{R}^{6\times 6}$ are the mass, damping, and stiffness diagonal matrices, respectively, which are used to impose specific second-order mass-spring-damper dynamics; $\tilde{{\bf h}}^{\mathcal R_{ed}}(k) = {\bf h}_d^{\mathcal R_{ed}}(k) - {\bf h}_e^{\mathcal R_{ed}}(k)$ is the wrench tracking error, defined as the difference between the desired wrench ${\bf h}_d^{\mathcal R_{ed}}(k)$ and the exerted one ${\bf h}_e^{\mathcal R_{ed}}(k)$, which is equal and opposite to the measured wrench.
We take ${\bf h}_d^{\mathcal R_{ed}}(k) = {\bf 0}, \forall k \in \mathbb N_0$ to make the robot behave compliantly with the environment.

\subsubsection{Free-hand force limitation strategy}\label{sec:free-hand-force-limitation-strategy}
In an admittance-controlled system, the interaction wrenches emerge from the combination of the admittance parameters, the mechanical properties of the environment, and the reference provided to the controller \cite{gao_combined_2023}. 
It follows that interaction forces can be limited through a suitable reference ${\bf x}_{\mathcal R_{ed}}^{\mathcal R_e}$. Accurate limitation requires perfect knowledge of the environment that is particularly challenging to achieve, especially in a medical environment, in which the FR interacts with surfaces of varying stiffness and non-uniform geometry. Therefore, in the proposed strategy, the position reference
\begin{equation}
    {\bf p}_{\mathcal R_{ed}}^{\mathcal R_e}(k) = {\bf R}_{\mathcal R_b}^{\mathcal R_e}(k) {\bf p}_{\mathcal R_{ed}}^{\mathcal R_b}(k) - {\bf R}_{\mathcal R_b}^{\mathcal R_e}(k) {\bf p}_{\mathcal R_e}^{\mathcal R_b}(k),
\end{equation}
is dynamically scaled based solely on the contact forces:
\begin{equation}\label{eq:force-limitation-strategy}
    \underline{{\bf p}}_{\mathcal R_{ed}}^{\mathcal R_e}(k) = \frac{1}{1 + K_s \lVert {\bf f}_e^{\mathcal R_e}(k) \rVert} {\bf p}_{\mathcal R_{ed}}^{\mathcal R_e}(k),
\end{equation}
where $K_s \in \mathbb R_+$ is a proportional gain representing a trade-off between smoothness and responsiveness of the scaling action. 
Indeed, $K_s$ close to $0$ reduces sudden variations of $\underline{{\bf p}}_{\mathcal R_{ed}}^{\mathcal R_e}(k)$, even in presence of noisy force measures, but provides less force-limiting action. On the other hand, $K_s$ significantly greater than $0$ makes the scaling action more responsive, but may yield oscillations in the presence of noisy measurements. Therefore, to increase the system's robustness to noisy forces, exhibiting high-frequency variations, ${\bf f}_e^{\mathcal R_e}$ is filtered online as done for positions in \eqref{eq:moving-average-filter}. We note that the strategy in \eqref{eq:force-limitation-strategy} does not allow forces to be bounded below a predefined limit. However, it prevents contact forces from growing indefinitely, ensuring a limitation even in the transients of the interaction and for any orientation of the end-effector, without relying on the knowledge of the dental environment's complex geometry.

From a control theory perspective, it is appropriate to draw an analogy between the proposed strategy and passivity-based methods, such as TDPA \cite{artigas_time_2010}, as both rely on modulating the reference provided to the downstream controller through an adaptive damping mechanism. 
However, a substantial difference exists in terms of control objectives. 
While passivity-based approaches provide an event-triggered modulation of the reference in response to violations of the energy balance, with the explicit goal of guaranteeing system passivity, the proposed strategy implements a continuous, non-energizing, force-driven reference modulation aimed at reducing interaction forces, without relying on explicit energy estimation.
As a consequence, its attenuative and smoothly varying action on the reference, resulting from an appropriate choice of the gain $K_s$ and of the force filtering window, does not inject energy into the system, nor does it alter the underlying dynamics, in contrast to state-of-the-art variable admittance–based approaches \cite{gao_combined_2023, cho_sensorless_2023}, which explicitly modify the controller dynamics and typically rely on explicit knowledge of the environment \cite{roveda_sensorless_2022}.
As a result, the stability properties of the admittance-controlled interaction remain unaffected.
Moreover, as discussed above, passivity-based control addresses an orthogonal objective to the proposed approach, and can therefore be combined with the eye–hand coordination and force limitation strategies to guarantee stability even in the presence of considerable communication delays.
We consider this scenario in Section \ref{sec:force-limitation-assessment}.

\subsubsection{Safety mechanisms and emergency stop protocols}\label{sec:safety-mechanisms}
From \eqref{eq:admittance-model} and \eqref{eq:force-limitation-strategy}, the compliant pose ${\bf x}_{\mathcal R_c}^{\mathcal R_b}(k)$ to be commanded to the robot is obtained. Specifically,
\begin{align}
    {\bf p}_{\mathcal R_c}^{\mathcal R_b}(k) & = \underline{{\bf p}}_{\mathcal R_{ed}}^{\mathcal R_b}(k) + {\bf R}_{\mathcal R_{ed}}^{\mathcal R_b}(k) {\bf p}_{\mathcal R_c}^{\mathcal R_{ed}}(k), \\
    {\bf q}_{\mathcal R_c}^{\mathcal R_b}(k) & = {\bf q}_{\mathcal R_{ed}}^{\mathcal R_b}(k) {\bf q}_{\mathcal R_c}^{\mathcal R_{ed}}(k),
\end{align}
where ${\bf x}^{\mathcal R_{ed}}_{\mathcal R_c}$, taking the semantics of a pose error, is obtained by time-integrating $\dot{\bf v}^{\mathcal R_{ed}}_{\mathcal R_c}$ from \eqref{eq:admittance-model} twice.
In addition to its compliant interaction properties and the embedded force limitation strategy, the controller incorporates two safety mechanisms and implements an emergency stop protocol. 
Regarding the safety mechanisms, the first ensures the stability of the FR in the absence of updated references from the EHCC ${\bf x}_{\mathcal R_{ed}}^{\mathcal R_b}(k)$, such as in the event of a communication loss, by continuing to track the last valid one ${\bf x}_{\mathcal R_{ed}}^{\mathcal R_b}(k-1)$. 
The second mechanism prevents abrupt motions of the FR by commanding ${\bf x}_{\mathcal R_{e}}^{\mathcal R_b}(k)$ whenever the generated command ${\bf x}_{\mathcal R_c}^{\mathcal R_b}(k)$ deviates excessively, either in translation or orientation, based on translational and rotational safety thresholds that can be configured by the operator according to the task requirements.
With regard to the emergency stop, it focuses on the monitoring of the interaction forces. 
Whenever $\lVert {\bf f}^{\mathcal R_e}\rVert(k)$ exceeds a predefined safety threshold $f_{th}$, the controller forwards ${\bf x}_{\mathcal R_{e}}^{\mathcal R_b}(k)$  as the new reference while continuing to compute the admittance control law. 
In this way, abrupt stopping is avoided, while still allowing the interaction forces to decrease progressively.

\subsection{Haptic feedback controller}\label{sec:haptic-feedback-controller}
In performing medical procedures, the operator relies heavily on the sense of touch. Therefore, in a HBTS providing realistic haptic sensations, it becomes critical to the success of the procedures. In impedance systems, the interaction is captured by both the contact force ${\bf f}_e^{\mathcal R_e}$ and the position error $\tilde{\bf p}^{\mathcal R_e}$, the latter being characterized by slower transients than the former, which exhibits a step dynamics. 
As proposed in \cite{pagliara_safe_2024}, replacing ${\bf f}_e^{\mathcal R_e}$ with $\tilde{\bf p}^{\mathcal R_e} = {\bf p}^{\mathcal R_e}_{\mathcal R_{ed}} - {\bf p}_{\mathcal R_c}^{\mathcal R_e}$ to generate the force feedback makes each contact state transparent to the operator, preventing them from having to deal with sudden force changes they are not used to from previous training and fieldworks. 
Therefore, the proposed HFC defines a second (virtual) impedance model, generating the force feedback from the motion error of the admittance controller:
\begin{equation}\label{eq:virtual-feedback}
{\bf f}_h^{\mathcal H_b}(k) = {\bf K}_{h} \tilde{{\bf p}}^{\mathcal H_b}(k) + {\bf D}_{h} \dot{\tilde{{\bf p}}}^{\mathcal H_b}(k),
\end{equation}
where ${\bf K}_{h} \in \mathbb{R}^{3\times 3}$ and ${\bf D}_{h} \in \mathbb{R}^{3\times 3}$ respectively are virtual stiffness and damping diagonal matrices, where
\begin{equation}
    \begin{bmatrix}
        \tilde{{\bf p}}^{\mathcal H_b}(k) \\
        \dot{\tilde{{\bf p}}}^{\mathcal H_b}(k)
    \end{bmatrix} =
     \Bigl( {\bf I}_2 \otimes \bigl(
        {\bf R}_{\mathcal H_e}^{\mathcal H_b}(k)
        {\bf R}_{\mathcal R_e}^{\mathcal H_e}(k)
     \bigr)
     \Bigr)
     \begin{bmatrix}
         \tilde{{\bf p}}^{\mathcal R_e}(k) \\
         \dot{\tilde{{\bf p}}}^{\mathcal R_e}(k)
     \end{bmatrix}.
\end{equation}
We note that, as in \eqref{eq:FR-twist}, the rotation matrix ${\bf R}_{\bf z}^{\mathcal R_e}(k)$ (or equivalently ${\bf R}_{\bf z}^{\mathcal H_e}(k)$) in ${\bf R}_{\mathcal H_e}^{\mathcal R_e}(k)$ is used to provide the force feedback correspondence with the interaction observed in the camera view. 
This approach is particularly suited to the considered application, as it guarantees stability while providing the operator with a natural and intuitive tactile experience that closely resembles manual execution, especially in light of Remark \ref{rem:sense-of-touch}. 
Within this framework, the parameters ${\bf K}_h$ and ${\bf D}_h$ of the virtual system in \eqref{eq:virtual-feedback} are tuned to shape the perceived interaction, with the stiffness matrix ${\bf K}_h$ governing the transient response and the damping matrix ${\bf D}_h$ contributing to vibration attenuation.
Moreover, the use of a second virtual impedance model to generate the force feedback moderates the influence of transient force spikes associated with sudden contacts, while preserving informative and natural haptic feedback.
We envisage the possibility that the generated feedback may exceed the range of forces that the HD device can exert. 
Therefore, the HFC saturates the norm of the force feedback $\lVert{\bf f}_h^{\mathcal H_b}(k)\rVert$ to the maximum exertable force $f_{\text{max}}^{\mathcal H_b} \in \mathbb R_+$:
\begin{equation}
    \underline{\mathbf{f}}_h^{\mathcal{H}_b}(k) =
    \begin{cases}
        \mathbf{f}_h^{\mathcal{H}_b}(k), & \text{if}~ \lVert \mathbf{f}_h^{\mathcal{H}_b}(k)\rVert < f_{\text{max}}^{\mathcal{H}_b};\\
        f_{\text{max}}^{\mathcal{H}_b}\dfrac{\mathbf{f}_h^{\mathcal{H}_b}(k)}{\lVert \mathbf{f}_h^{\mathcal{H}_b}(k)\rVert}, & \text{otherwise}.
    \end{cases}
\end{equation}
Finally, given the slow dynamics of $\tilde{{\bf p}}^{\mathcal R_e}$, the operator may perceive residual force even when the robot is no longer interacting. 
To address this, we introduce a force dead-band $f_{db} \in \mathbb R_+$ to prevent the HD from reacting to such residuals:
\begin{equation}
    \overline{{\bf f}}_h^{\mathcal H_b}(k) =
    \begin{cases}
        \underline{{\bf f}}_h^{\mathcal H_b}(k), &\text{if}~ \bigl\lVert {\bf f}_e^{\mathcal R_e}(k) \bigr\rVert > f_{db};\\
        {\bf 0}, &\text{otherwise}.
    \end{cases}
\end{equation}

\subsection{Integrated mechanical end-effector}\label{sec:integrated-mechanical-end-effector}
In order to successfully perform any free-hand dental procedure, the operator relies on three basic factors: (a) the procedure-specific clinical tool; (b) the surgical field view, which may require external illumination and tools to show obstructed or poorly lit areas, e.g., a mirror; (c) the sense of touch to ensure the right pressure to be exerted on the area of interest, in order to achieve a safe and effective procedure. In this work, we design an integrated mechanical end-effector that can provide all the tools and sensory capabilities required to perform free-hand dental procedures.
\begin{figure}
    \centering
    \includegraphics[width=\columnwidth]{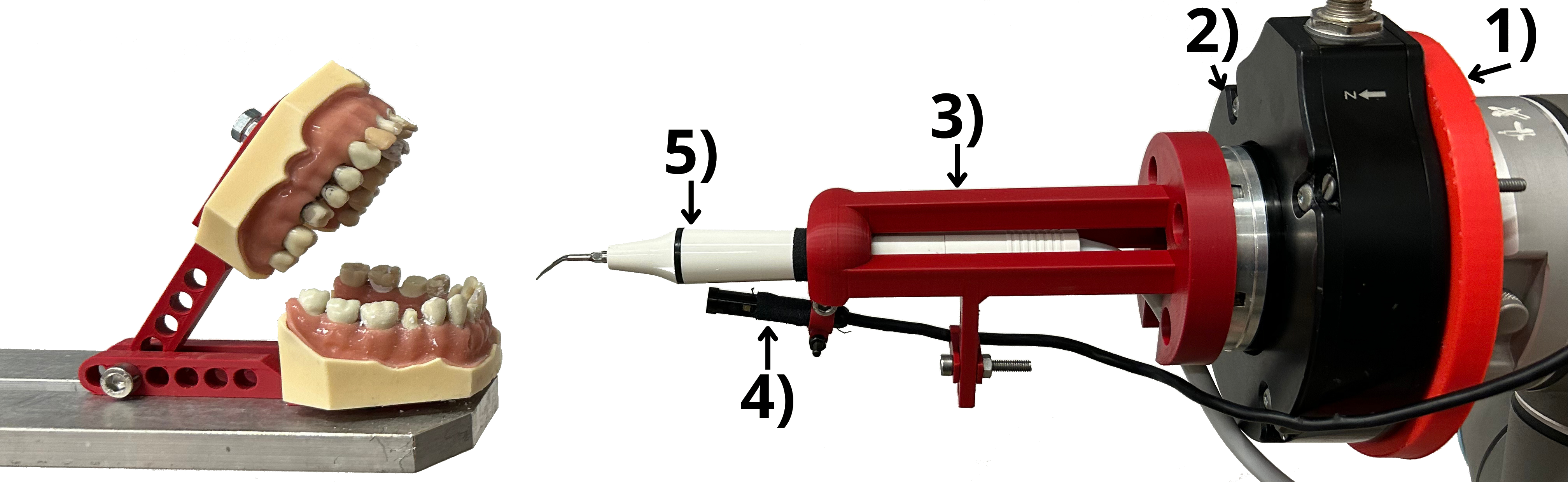}
    \caption{Integrated mechanical end-effector for free-hand procedures. Visible in the left side of the figure is also one of the dental models used in the experiments.}
    \label{fig:integrated_mechanical_end_effector}
\end{figure}

With reference to Figure \ref{fig:integrated_mechanical_end_effector}, the designed end-effector consists of five main components: (1) a
\textit{3D-printed mechanical interface} connecting the robot's flange with the force/torque sensor; (2) a \textit{force/torque sensor} measuring contact forces; (3) a \textit{3D-printed tool socket} supporting the clinical tool and the endoscopic camera: the interface consists of a rigid structure to enable proper transmission of forces from the tool to the sensor, a circular hole, adjustable via a bolt-and-nut mechanism, to accommodate tools of different sizes, and a movable camera-support ring, allowing for adjustments of the view and light beam angle; (4) an \textit{high definition endoscopic camera}, equipped with LED illumination to ensure good visibility even in poorly lit areas; (5) a \textit{clinical tool} for performing the procedure of interest: the tool is placed in the supporting socket after being wrapped with friction tape to ensure good mechanical compliance, emulating the skin friction between the operator's hand and the tool; the tool used for the chosen use case is a scaler.

Although the current implementation does not fulfill yet all clinical requirements, such a design is compatible with future clinical adoption, as the end-effector is inherently compatible with standard sterilization and infection-control procedures commonly adopted in dentistry.
Indeed, clinical instruments are intrinsically designed to withstand autoclave sterilization, and the literature reports the availability of sterilizable endoscopic cameras \cite{boese_endoscopic_2022}, as well as autoclavable or low-temperature–sterilizable materials, including medical-grade 3D-printed polymers, which would enable full sterilization of the entire assembly \cite{rynio_effects_2022}. 
Moreover, the tool socket has been designed with smooth external surfaces and a reduced number of crevices, thereby facilitating cleaning and reprocessing.
In addition, the socket enables rapid tool replacement by simple extraction and insertion of the clinical instrument, without requiring disassembly of the end-effector or disconnection of sensing and vision components.
Components upstream of the end-effector, such as the force/torque sensor and the robot flange, which are not compatible with autoclave sterilization, can be isolated using sterile drapes or disposable sterile barriers, as commonly adopted in medical robotic systems \cite{van_der_schans_da_2020}. 
With respect to aerosol management, a critical concern in dental procedures such as SRP, the proposed design does not interfere with conventional mitigation strategies. 
In a clinical deployment, aerosol generation could be managed through standard high-volume evacuation systems and saliva ejectors, without requiring any modification to the design.

\section{Technical validation}\label{sec:technical-validation}
To evaluate the performance of the proposed FH-HBTS, we conduct a set of experiments including static force application (Section \ref{sec:force-limitation-assessment}), dynamic dental scaling motions (provided in the supplementary materials), and a free-space experiment (Section \ref{sec:eye-hand-coordination-assessment}). 
During the experiments, the robot is out of the operator's view, who relies solely on the visual feedback from the camera and the force feedback from the HD. We select three target teeth: the lower right first premolar, the upper right central incisor, and the upper right second premolar. 
The chosen targets are significant as each requires a different tool orientation, making the use of eye-hand coordination essential.

\subsection{Experimental setup}
The experimental setup involves the following equipment and materials: (I) a leader control station featuring a 6-degrees-of-freedom (DOF) force-feedback HD\footnote{\url{https://www.3dsystems.com/haptics-devices/touch}} and a workstation: the latter is a general purpose computer executing the leader controller loop, and communicating with the FR over an Ethernet network, where delays and packet losses are negligible; (II) a FR consisting of the 6-DOF UR10 arm\footnote{\url{https://www.universal-robots.com/}}, equipped with the integrated mechanical end-effector of Section \ref{sec:integrated-mechanical-end-effector}, where the force/torque sensor is the Robotiq FT150\footnote{\url{https://robotiq.com}}; (III) a high-strength resin dental model and a 3D-printed dental phantom, both life-sized models of a human mouth.
The former is used for the experiments described in this section, while the latter only appears in the accompanying video.

\subsection{Force limitation assessment}\label{sec:force-limitation-assessment}
To assess the effectiveness of the proposed force limitation strategy, we perform $9$ interaction experiments, with an average duration of $\sim\SI{20}{\second}$ each, 3 for each target tooth, without force limitation (Scenario A), $9$ more with force limitation (Scenario B) and $9$ with a passivity-based architecture (Scenario C), specifically TDPA-HD, where admittance control replaces Cartesian impedance control adopted in \cite{panzirsch_exploring_2022}. 
In the latter scenario, the force limitation is intentionally kept disabled in order to isolate the effects of TDPA-HD.
For each scenario, we record the data sequence $\bigl\{a(k), b(k)\bigr\}_{k=1}^{N}$, sampling at a frequency of \SI{125}{Hz}, where we set the force norm $\left\lVert {\bf f}_e^{\mathcal R_e}(k) \right \rVert$ as $a(k) \in \mathcal A \subseteq \mathbb R_+$, and the commanded virtual penetration $\bigl\lVert {\bf p}_{\mathcal R_{ed}}^{\mathcal R_e}(k)\bigr\rVert$ as $b(k) \in \mathcal B = \{b \in \mathbb R \vert b_{\text{min}} = \SI{0}{\meter} \leq b \leq b_{\text{max}} = \SI{0.007}{\meter}\}$.
We discretize $\mathcal B$ with the step $b_{\text{step}} = \SI{0.0001}{\meter}$, selected based on the repeatability of the FR, while $b_{\text{max}}$ is the maximum virtual penetration observed across all experiments.

\begin{figure}
    \centering
    \includegraphics[width=1\columnwidth]{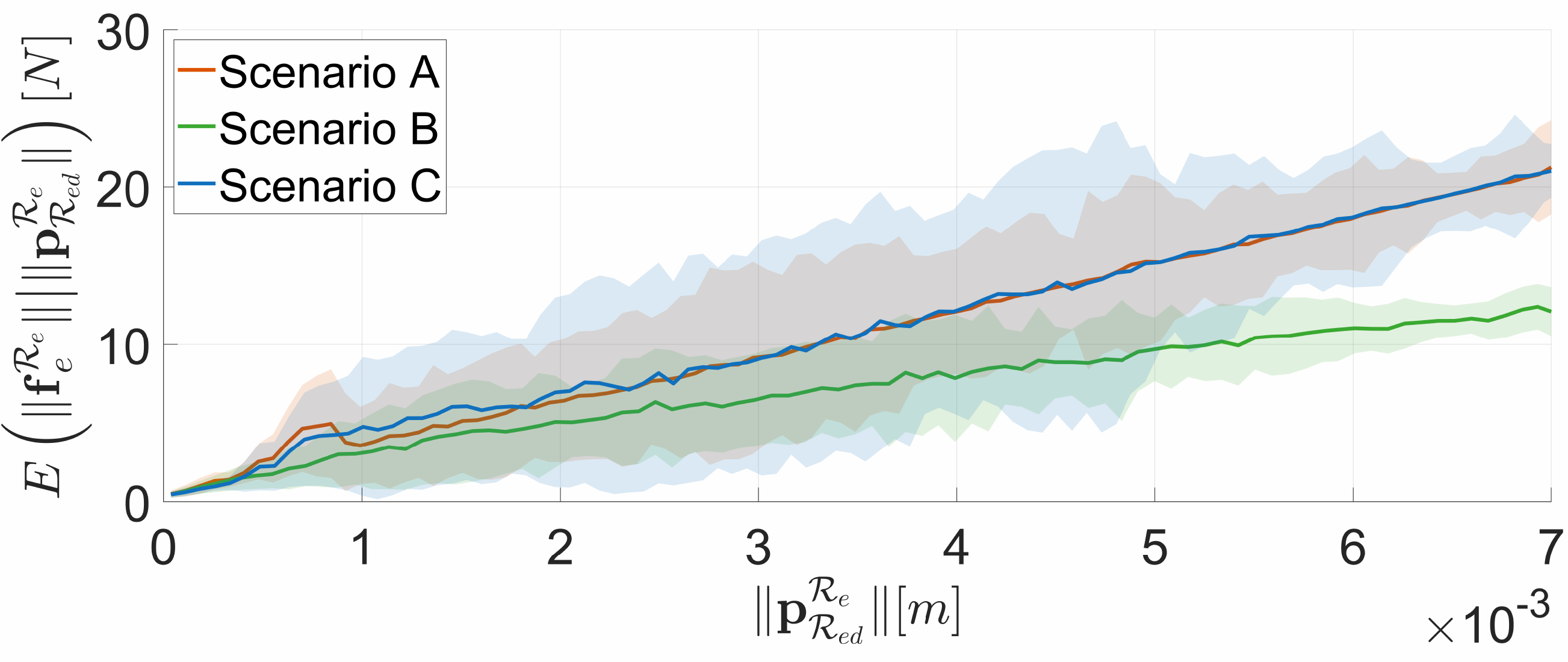}
    \caption{Comparison between the conditional expectation of the interaction force in all three scenarios.}
    \label{fig:conditional_expectation}
\end{figure}

By binning data in $\mathcal B$, following \cite{steyer_probability_2017}, we estimate the conditional expectation of the exerted force $E\bigl(\lVert {\bf f}_e^{\mathcal R_e}\rVert \big\vert \lVert {\bf p}_{\mathcal R_{ed}}^{\mathcal R_e}\rVert\bigr)$ and the conditional standard deviation, obtained as the square root of the conditional variance $\text{Var}\bigl(\lVert {\bf f}_e^{\mathcal R_e} \rVert \big\vert \lVert {\bf p}_{\mathcal R_{ed}}^{\mathcal R_e}\rVert\bigr)$, in all three scenarios.
Visual inspection of Figure \ref{fig:conditional_expectation} showcases a significant reduction in the expected interaction force in Scenario B, given the same commanded virtual penetration.
Notably, TDPA-HD \cite{panzirsch_exploring_2022} (Scenario C) cannot reduce the interaction force in quasi-static contacts due to the energy being approximately zero.

To evaluate the effectiveness of the force limitation strategy, we compare $E\bigl(\lVert {\bf f}_e^{\mathcal R_e}\rVert \big\vert \lVert {\bf p}_{\mathcal R_{ed}}^{\mathcal R_e}\rVert\bigr)$ and $\text{Var}\bigl(\lVert {\bf f}_e^{\mathcal R_e} \rVert \big\vert \lVert {\bf p}_{\mathcal R_{ed}}^{\mathcal R_e}\rVert\bigr)$ in all three scenarios.
We assess differences in conditional variances through Levene's and Bartlett's tests in the comparisons involving Scenario B, i.e., B--A and B--C. 
Both tests indicate significantly different variances ($p \approx 0$), showing that the force limitation strategy affects the dispersion of the force norm $\bigl\lVert {\bf f}_e^{\mathcal R_e} (k)\bigr\rVert$ with respect to virtual penetration $\bigl\lVert {\bf p}_{\mathcal R{ed}}^{\mathcal R_e}(k) \bigr\rVert$.
Since Levene's test suggests a significant variance difference, we apply Welch's t-test, which is more robust in the case of unequal variances. 
In the B--A comparison, the test yields a t-statistic $t = -4.52$ with $p \approx 0$, confirming a statistically significant force reduction in Scenario B.
Beyond statistical significance, the magnitude of this effect is notable: the comparison yields a Cohen's $d = -0.63$, indicating a significant force reduction when the force limitation strategy is enabled.
The corresponding confidence interval $[-0.91,-0.36]$ does not include zero, further supporting the robustness and consistency of the observed effect across the explored range of virtual penetration.

An analogous analysis is conducted for the B-C comparison. Also in this case, Welch's t-test confirms a statistically significant reduction of the normal force with respect to Scenario C ($p \approx 0$), indicating that the force limitation strategy yields lower contact forces than the passivity-based control architecture. The observed effect is comparable in magnitude to that of the B-A comparison (Cohen's $d = -0.68$ and confidence interval $[-0.97,-0.41]$), further supporting the robustness of the proposed approach across different control architectures.

\subsection{Eye-hand coordination assessment}\label{sec:eye-hand-coordination-assessment}
\begin{figure}
    \centering
    \subfloat{\includegraphics[width=\columnwidth, trim=-50pt 0pt 0pt 0pt]{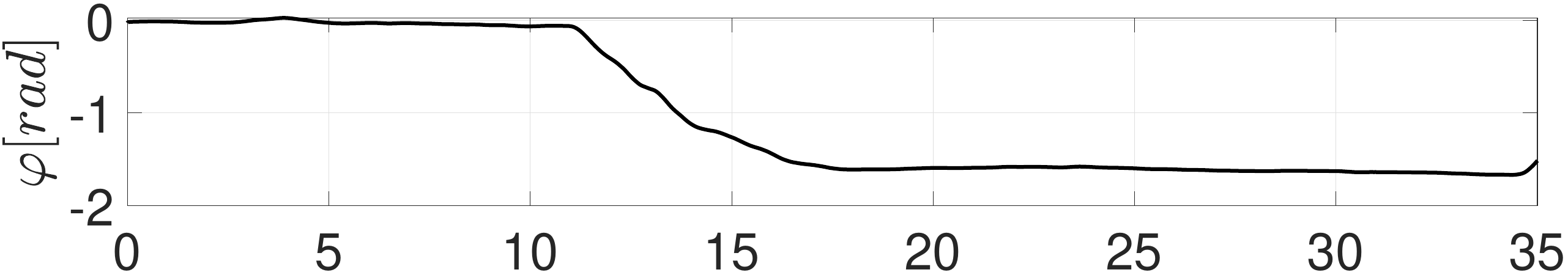}}
    \\
     \subfloat{\includegraphics[width=\columnwidth]{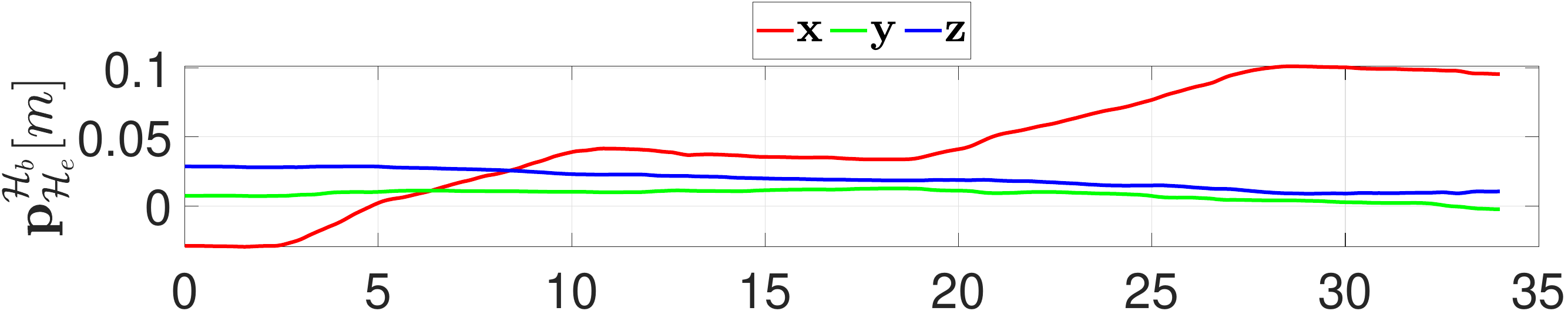}}
     \\
     \subfloat{\includegraphics[width=\columnwidth]{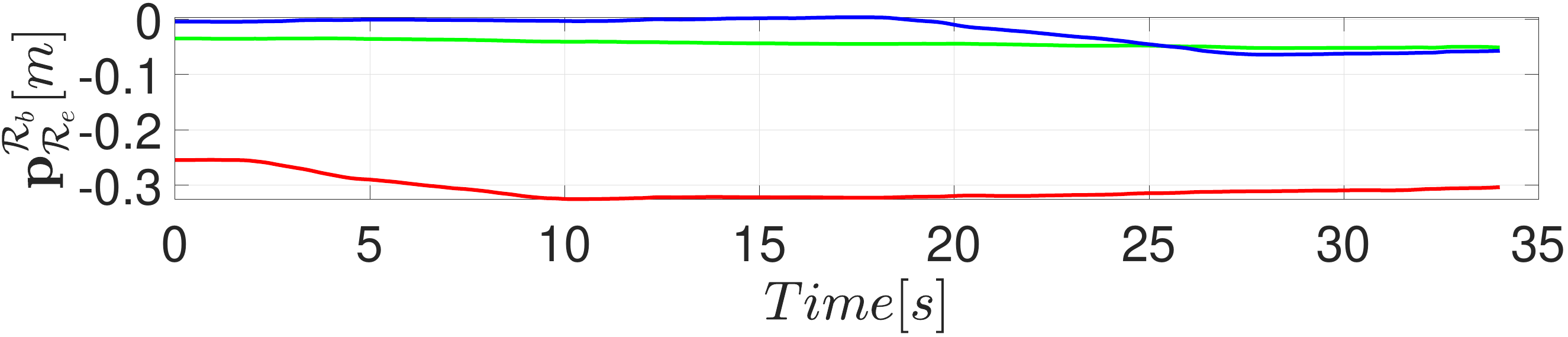}}
    \caption{Eye-hand coordination assessment.}
    \label{fig:eye-hand-coordination-validation}
\end{figure}

To perform a formal assessment of eye-hand coordination, we execute a teleoperated task in the free space. The operator performs a translatory motion with the HD, simultaneously rotating the viewing angle by rotating the HD's stylus by an angle $\varphi$ about the $\bf z$ axis of $\mathcal H_e$. The validation is considered completed if the motion transferred to the robot varies as the viewing angle does. Figure \ref{fig:eye-hand-coordination-validation} shows the $\varphi$ angle, the position of HD's stylus and the position of the FR, respectively.

Results showcase that, as long as the angle $\varphi = \SI{0}{\radian}$ (from $\SI{0}{\second}$ to $\SI{10}{\second}$), a motion along the $x$ axis of $\mathcal H_b$ is mapped into a motion along the $x$ axis of $\mathcal R_b$. When the angle turns into $\varphi = - \pi/\SI{2}{\radian}$, a motion along the $x$ axis of $\mathcal H_b$ is mapped into a motion along the $z$ axis of $\mathcal R_b$ (from $\SI{20}{\second}$ to $\SI{30}{\second}$).

These results confirm that the mapping of motion between HD space and FR space is affected by the rotation of the visual field. However, they do not provide any suggestion regarding the effectiveness of eye-hand coordination for the human operator. Therefore, in the accompanying video we provide a comprehensive qualitative assessment of the latter. 

\section{In-vitro pre-clinical validation}\label{sec:in-vitro-clinical-validation}
To validate the feasibility of the FH-HBTS in performing free-hand dental procedures, we conduct an in-vitro experimental user study involving dental professionals and replicating a SRP procedure under more representative clinical conditions.
With respect to real clinical practice, the procedure is simplified in that the presence of fluids and cheek occlusions are neglected.
The procedure is carried out on the dental model, coated with an acrylic polymer–based suspension containing titanium dioxide to simulate artificial calculus, using a fully functional ultrasonic scaler activated via a foot pedal at the leader control station.

A total of 20 participants with varying levels of experience in the dental field are recruited: (\textit{Group 1}) dental students who had performed at least 20 in-vivo dental scaling procedures $(30\%)$; (\textit{Group 2}) dentists with less than 5 years of professional experience $(35\%)$; (\textit{Group 3}) dentists with more than 5 years of professional experience $(35\%)$. 
Before the experimental phase, all participants complete a structured training session aimed at ensuring competent use of the system, during which they practice approaching the dental surface, maintaining stable contact, and performing preliminary calculus removal.

After training, participants perform artificial calculus removal on selected tooth surfaces: lower left central incisor, lower left second molar, lower right first molar, relying on both visual and haptic feedback. For each trial, each surface is uniformly coated with a thin layer of artificial calculus.
The procedure is considered successful if the participant is satisfied with the outcome and none of the safety mechanisms described in Section \ref{sec:safety-mechanisms} are triggered.
A single representative trial, illustrating the overall experimental procedure, is provided in the accompanying video.

At the end of the experimental session, each participant completes a questionnaire to provide feedback on their experience with the system.
These subjective responses are complemented by objective measurements collected during task execution and by independent expert assessments, providing a comprehensive evaluation of the experimental procedure, as described in the following subsection.

\subsection{Data collection and analysis}
To comprehensively evaluate system performance, operator experience, and operational safety, a set of objective and subjective outcome measures is collected and analyzed. 
These include indicators of clinical efficacy, task efficiency, interaction safety, and perceived workload and usability. 
Differences across participant experience levels are assessed for all outcome measures using Kruskal–Wallis tests.

System efficacy is evaluated by an independent expert dentist, blind to the participant's experience group, who assesses calculus removal using a six-point ordinal scale ranging from no removal to complete removal. 
Task efficiency is quantified through the median execution time per dental surface, while interaction safety is assessed by analyzing the magnitude of the interaction forces exerted during the procedure. Both metrics are compared with reference values reported in the literature for manual execution. 
Finally, system usability is evaluated through subjective measures, including the NASA-Task Load Index (NASA-TLX) questionnaire for perceived workload, the System Usability Scale (SUS) questionnaire for usability, and additional 5-point Likert-scale questions to further characterize the operator interaction.
The complete questionnaire is reported in the supplementary materials.

\subsection{Results}
All participants were able to successfully complete the entire experimental procedure.
Regarding system efficacy, the mean cleanness score per participant, averaged over the three teeth, ranges from a minimum of $3.66$ to a maximum of $4.28$, indicating a generally high and homogeneous level of task execution.
No statistically significant differences between groups are observed, as confirmed by the Kruskal–Wallis test conducted at a significance level of $p = 0.05$.

Regarding execution time, the duration of the procedure per tooth showcases substantial variability, ranging from \SI{85}{\second} to \SI{212}{\second}.
The median execution times are \SI{153.5}{\second} for Group 1, \SI{162.0}{\second} for Group 2, and \SI{147.5}{\second} for Group 3.
Reference values reported in \cite{ruppert_vivo_2002} for manual performance of the same procedure indicate a median execution time of \SI{115.5}{\second}.
Although the teleoperated procedure is therefore slower than manual execution, participants exhibited an average reduction of $20.4 \%$ in completion time in the second trial, followed by a further $23.2\%$ reduction in the third trial.
This trend reflects a clear learning-curve effect, indicating that operators progressively familiarize themselves with the system and improve their ability to control the remote tool. 
This suggests that, with more extensive training, teleoperated performance could gradually approach manual execution levels.

In terms of interaction forces, during the experimental trial, the force magnitude results in an overall mean of $1.78 \pm 0.19\,\mathrm{N}$ for Group 1, $1.86 \pm 0.34\,\mathrm{N}$ for Group 2, and $1.96 \pm 0.32\,\mathrm{N}$ for Group 3, with average maximum peaks of $9.44 \pm 3.09\,\mathrm{N}$, $9.80 \pm 4.7\,\mathrm{N}$, and $11.47 \pm 4.41\,\mathrm{N}$, respectively. 
When comparing groups, Kruskal-Wallis tests find no statistically significant differences between experience levels.
The comparison with reference values reported in the literature for dentists performing the procedure free-hand highlights a clear discrepancy. 
In particular, \cite{stutzer_-vitro_2024} reports an average interaction force of $0.28 \pm \SI{0.33}{\newton}$, with average maximum peaks of approximately $\SI{0.60}{\newton}$. 
These values are close to the tissue safety thresholds reported in the literature for dental procedures, commonly identified at around $\SI{0.5}{\newton}$ \cite{flemmig_effect_1998}.
Notably, these reference values are also compatible with those measured in a dedicated free-hand SRP procedure we conduct on the dental model mounted on the force/torque sensor to directly measure interaction forces.
Further details on this experiment are provided in the supplementary materials.
This difference reflects an inherent limitation of the system, which is primarily attributable to the mechanical characteristics of the adopted industrial robot and to the resolution of the force/torque sensor, not specifically designed to accurately manage the low-magnitude forces typical of dental interactions. 
Despite the higher measured forces, no superficial damage was observed on the dental model, suggesting that the system remains safe under the adopted experimental conditions. 
However, these results should be regarded as scalable, as the interaction forces are strongly influenced by the employed hardware setup and are expected to decrease when adopting an engineered system tailored to the target application, particularly in terms of mechanical compliance and force/torque sensing resolution.

The results of the Kruskal–Wallis tests on questionnaire responses are reported in Table~\ref{tab:questionnaire}.
No statistically significant differences emerge across any NASA-TLX dimension, SUS score, or 5-point Likert item ($p>0.05$), indicating that subjective evaluations are consistent across experience levels.
Across all groups, the NASA-TLX results indicate a consistently moderate perceived workload, with contained mental, physical, and temporal demands, limited variability in performance ratings, and generally low effort and frustration.
According to the interpretation bands derived from a cross-domain meta-analysis of raw NASA-TLX scores \cite{Grier_2015}, the obtained Overall values (mean $\in [25.71, 26.94]$) fall within the low-workload range.
Overall, the task is perceived as manageable and not overly demanding.
System usability is evaluated positively, with SUS scores falling within the good-usability range \cite{bangor_determining_2009}, confirming that the system is considered intuitive and easy to use.
Likert-scale evaluations further highlight high ratings for interaction-related aspects, including motion and force correspondence, first contact and prolonged interaction perception, and the naturalness of the haptic feedback (mean $\in [3.43, 4.33]$), suggesting coherent and supportive sensory feedback. 
Ergonomics is also positively evaluated (mean $\in [3.71, 4.17]$).
This Likert item was explicitly anchored relative to the participant's own experience of manual execution, so that a score of $3$ denotes the neutral midpoint of the scale, i.e., perceived parity between the teleoperated and manual conditions, while scores above (resp.\ below) $3$ denote a perceived improvement (resp.\ deterioration); this reference value is intrinsic to the anchoring of the scale (see the complete questionnaire in the supplementary materials) rather than derived from an external source.
The obtained scores, all above this neutral point, indicate an improved ergonomic and postural comfort during teleoperation, consistent with the contained Physical Demand reported in the NASA-TLX (mean $\in [16.67, 26.67]$).
Fidelity (mean $\in [3.00, 3.33]$) achieves only moderately positive scores, likely due to limitations in depth perception (mean $\in [1.71, 2.33]$), preventing accurate situational awareness.

\begin{table*}
\centering
\begin{tabular}{|l|c|c|c|cc|} \hline
 & \textbf{Group 1} & \textbf{Group 2} & \textbf{Group 3} & \multicolumn{2}{c|}{\textbf{K–W test}} \\ \hline 
 & Mean $\pm$ STD & Mean $\pm$ STD & Mean $\pm$ STD & $\chi^2$ & p \\ \hline
\multicolumn{6}{|l|}{\textbf{NASA-TLX}} \\ \hline
Mental Demand    & $30.00 \pm 27.57$ & $51.43 \pm 15.74$ & $36.67 \pm 23.38$ & 2.91 & 0.23 \\ \hline
Physical Demand  & $26.67 \pm 24.22$ & $20.00 \pm 11.55$ & $16.67 \pm 19.66$ & 0.70 & 0.70 \\ \hline
Temporal Demand  & $36.67 \pm 23.38$ & $42.86 \pm 13.80$ & $36.67 \pm 15.06$ & 0.52 & 0.77 \\ \hline
Performance      & $43.33 \pm  8.17$ & $37.14 \pm 13.80$ & $33.33 \pm 10.33$ & 2.42 & 0.30 \\ \hline
Effort           & $16.67 \pm 19.66$ & $ 8.57 \pm 10.69$ & $23.33 \pm 15.06$ & 2.79 & 0.25 \\ \hline
Frustration      & $30.00 \pm 27.57$ & $28.57 \pm 27.95$ & $33.33 \pm 16.33$ & 0.11 & 0.94 \\ \hline
Overall          & $26.19 \pm 18.65$ & $26.94 \pm 13.36$ & $25.71 \pm 14.26$ & 2.00 & 0.37 \\ \hline
\hhline{|=|=|=|=|==|}
\textbf{SUS} & $77.92 \pm 13.46$ & $81.07 \pm 9.11$ & $78.75 \pm 10.34$ & 0.51 & 0.78 \\
\hhline{|=|=|=|=|==|}
\multicolumn{6}{|l|}{\textbf{5-point Likert Scale}}\\ \hline
Ergonomics            & $3.83 \pm 0.75$ & $3.71 \pm 0.76$ & $4.17 \pm 0.98$ & 0.95 & 0.62 \\ \hline
Fidelity              & $3.17 \pm 1.33$ & $3.00 \pm 0.58$ & $3.33 \pm 0.52$ & 0.87 & 0.65 \\ \hline
Motion correspondence & $4.17 \pm 0.41$ & $4.00 \pm 0.82$ & $4.33 \pm 0.82$ & 0.78 & 0.68 \\ \hline
Force correspondence  & $4.00 \pm 0.89$ & $4.14 \pm 0.38$ & $4.17 \pm 0.75$ & 0.17 & 0.92 \\ \hline
Depth perception      & $2.17 \pm 0.98$ & $1.71 \pm 0.76$ & $2.33 \pm 0.52$ & 2.19 & 0.33 \\ \hline
Interaction events    & $3.83 \pm 0.98$ & $3.43 \pm 0.98$ & $4.00 \pm 1.10$ & 1.30 & 0.52 \\ \hline
Interaction effects   & $3.83 \pm 0.98$ & $3.43 \pm 0.79$ & $4.17 \pm 0.75$ & 2.73 & 0.26 \\ \hline
Feedback naturalness  & $4.00 \pm 0.89$ & $3.71 \pm 0.95$ & $3.83 \pm 1.17$ & 0.33 & 0.85 \\ \hline
\end{tabular}
\caption{Summary of the NASA-TLX and 5-point Likert scale questionnaire responses and SUS scores.
NASA-TLX subscales are reported on their native 0--100 scale; the Overall row is the average of the six subscales.
Likert-scale items are reported on a 1--5 scale.
For the Ergonomics item, the scale was anchored relative to the participant's experience of manual execution, so that $3$ denotes the neutral midpoint (perceived parity with manual practice), while scores above/below $3$ denote a perceived improvement/deterioration relative to manual practice.}
\label{tab:questionnaire}
\end{table*}

The pattern of subjective outcomes is consistent with the design rationale underlying the proposed FH-HBTS, and can be related to the specific technical solutions introduced in Section~\ref{sec:methodology}.
The contained Effort (mean $\in [8.57, 23.33]$) and Frustration (mean $\in [28.57, 33.33]$) ratings are compatible with the role of the human-tremor filtering stage (Section~\ref{sec:human-tremors-filtering}) and of the dynamic velocity-scaling mechanism embedded in the EHCC (Section~\ref{sec:eye-hand-coordination}), both explicitly aimed at reducing the cognitive effort required to control the FR accurately.
The high scores obtained for motion and force correspondence (mean $\in [4.00, 4.33]$) are coherent with the effectiveness of the view-consistent eye–hand coordination mapping (Section~\ref{sec:eye-hand-coordination}) and of the force-feedback correspondence strategy implemented by the HFC (Section~\ref{sec:haptic-feedback-controller}), both specifically designed to preserve, in the camera view, a natural correspondence between the operator's actions and perceptions and the FR's motion and interaction.
Conversely, the comparatively lower Fidelity (mean $\in [3.00, 3.33]$) and, especially, Depth perception (mean $\in [1.71, 2.33]$) scores can be attributed to the monocular nature of the eye-in-hand endoscopic camera integrated in the mechanical end-effector (Section~\ref{sec:integrated-mechanical-end-effector}), which does not provide stereoscopic depth cues; this is an inherent limitation of monocular endoscopic vision, rather than of the eye-hand coordination or force-feedback control strategies, and motivates the exploration of stereo- or depth-sensing camera modules in future hardware iterations.
Finally, the favorable SUS scores and the improved Ergonomics ratings are consistent with the contained Physical Demand discussed above and reflect the removal of the operator from the direct, prolonged, and often awkward postures over the patient highlighted in Remark~\ref{rem:unstructured-environment}, which represents one of the central motivations for the proposed teleoperated approach.

\section{Conclusions and Future Perspectives}\label{sec:discussion-and-conclusions}

This paper proposed a FH-HBTS for free-hand dental procedures. The system features visual and haptic feedback, that are appropriately combined and coordinated to allow for a natural teleoperation experience. To improve the visibility of the intervention site, we designed a dedicated end-effector accommodating an eye-in-hand endoscopic camera, and ensuring compatibility with existing clinical tools. We also improved the accuracy of the procedure by introducing human tremor filtering and dynamic scaling of velocities. Through eye-hand coordination and  TCP control, we enabled the operator to perceive the motion correspondence between their movements and those of the end-effector, observed in the camera view, preserving manual dexterity and expertise. Additionally, by appropriately transforming the virtualized force feedback, we also ensured the force correspondence, enabling the human operator to experience a highly natural sensory perception. Finally, we enabled safe interactions by scaling the motion references provided to the admittance controller. The proposed strategy allows for accurate force limitation in any contact state and for any orientation of the interaction tool, without requiring any prior modeling of the intervention site.

We believe that our FH-HBTS represents a crucial step towards a broader adoption of robotics in dentistry,  as the collected in-vitro evidence suggests that accurate treatments can be performed, while preserving the operator's cognitive and technical abilities. With respect to manual procedures, our system allows improving operator's ergonomics, tracking operator's performance, and perspectively enabling remote healthcare.
Such results are nonetheless to be read in light of the simplified setup adopted, which does not yet embed all the clinically relevant sources of complexity.

We emphasize that the proposed system allows collecting haptic and motion data. While this feature is crucial during the training phase of new operators for the evaluation of learning metrics, i.e\ learning curves \cite{zhuang_exploring_2024}, it is also central towards the deployment of data-driven control architectures that leverage demonstrations to synthesize autonomous control policies \cite{gagliardi_probabilistic_2022}. Finally, these data can also be used to learn cost functions \cite{garrabe_convex_2025}, and hence intentions, of experts fulfilling a given task. In the context of dental procedures, this means that data collected from experts executing the procedure can be used to train policies for autonomous robots.

With our future work we plan to extend the FH-HBTS with autonomous patient motion compensation to further enhance system's safety and flexibility, and to benchmark the performance against the da Vinci Surgical System\textsuperscript{TM}. 
In addition, future experimental testbeds will be designed to progressively incorporate clinically relevant sources of complexity that are not yet addressed in the present study, including constrained inter-proximal access, interaction with soft tissues, fluid presence, and limited workspace conditions. 
In this context, we plan to collect expert demonstrations for the development of (semi-)autonomous dental procedures.
Beyond algorithmic extensions, a key direction of future work concerns the evolution of the experimental platform itself, which will be redesigned towards a fully engineered, application-driven system, explicitly accounting for the mechanical, sensing, and ergonomic requirements of dental procedures, and addressing the hardware-related limitations identified in the present study.

\section*{Acknowledgments}

The authors would like to thank Antonio Langella for their assistance during the in-vitro pre-clinical validation and Sabatino Panico for supporting the EHCC implementation and end-to-end system validation, making this work possible.

\bibliographystyle{IEEEtran}
\bibliography{bibliography}

\end{document}